\documentclass[sigconf]{acmart}
\AtBeginDocument{%
  }

\usepackage{multirow} 
\usepackage{mathtools}
\usepackage{subcaption}

\begin{document}

\settopmatter{printacmref=false} 
\renewcommand\footnotetextcopyrightpermission[1]{} 
\pagestyle{plain}  
\pagestyle{empty} 

\title{Advancing Opinion Dynamics Modeling with Neural Diffusion-Convection-Reaction Equation}

\author{Chenghua Gong}
\affiliation{%
  \institution{University of Science and Technology of China}
  \city{Hefei}
  \country{China}
}
\email{gongchenghua@mail.ustc.edu.cn}

\author{Yihang Jiang}
\affiliation{%
  \institution{University of Science and Technology of China}
  \city{Hefei}
  \country{China}
}
\email{jiang_yihang@mail.ustc.edu.cn}

\author{Hao Li}
\affiliation{%
  \institution{Wuhan University}
  \city{Wuhan}
  \country{China}
}
\email{whulh@whu.edu.cn}

\author{Rui Sun}
\affiliation{%
  \institution{University of Science and Technology of China}
  \city{Hefei}
  \country{China}
}
\email{rrsun@mail.ustc.edu.cn}

\author{Juyuan Zhang}
\affiliation{%
  \institution{University of Science and Technology of China}
  \city{Hefei}
  \country{China}
}
\email{zhangjuyuan2020@mail.ustc.edu.cn}

\author{Tianjun Gu}
\affiliation{%
  \institution{East China Normal University}
  \city{Shanghai}
  \country{China}
}
\email{51275901043@stu.ecnu.edu.cn}

\author{Liming Pan}
\authornote{Corresponding author.}
\affiliation{%
  \institution{University of Science and Technology of China}
  \city{Hefei}
  \country{China}
}
\email{pan_liming@ustc.edu.cn}

\author{Linyuan L\"u}
\authornotemark[1]
\affiliation{%
  \institution{University of Science and Technology of China}
  \city{Hefei}
  \country{China}
}
\email{linyuan.lv@ustc.edu.cn}

\begin{abstract}
    Advanced opinion dynamics modeling is vital for deciphering social behavior, emphasizing its role in mitigating polarization and securing cyberspace.
    To synergize mechanistic interpretability with data-driven flexibility, recent studies have explored the integration of Physics-Informed Neural Networks (PINNs) for opinion modeling.
    Despite this promise, existing methods are tailored to incomplete priors, lacking a comprehensive physical system to integrate dynamics from local, global, and endogenous levels.
    Moreover, penalty-based constraints adopted in existing methods struggle to deeply encode physical priors, leading to optimization pathologies and discrepancy between latent representations and physical transparency.
    To this end, we offer a physical view to interpret opinion dynamics via Diffusion-Convection-Reaction (DCR) system inspired by interacting particle theory.
    Building upon the Neural ODEs, we define the neural opinion dynamics to coordinate neural networks with physical priors, and further present the \textsc{Opinn}, a physics-informed neural framework for opinion dynamics modeling.
    Evaluated on real-world and synthetic datasets, \textsc{Opinn} achieves state-of-the-art performance in opinion evolution forecasting, offering a promising paradigm for the nexus of cyber, physical, and social systems.
    Code is available at~\url{https://anonymous.4open.science/r/OPINN-964F}
\end{abstract}


\maketitle

\section{Introduction}
\begin{figure}[htbp]
    \centering 
        \includegraphics[width=0.46\textwidth]{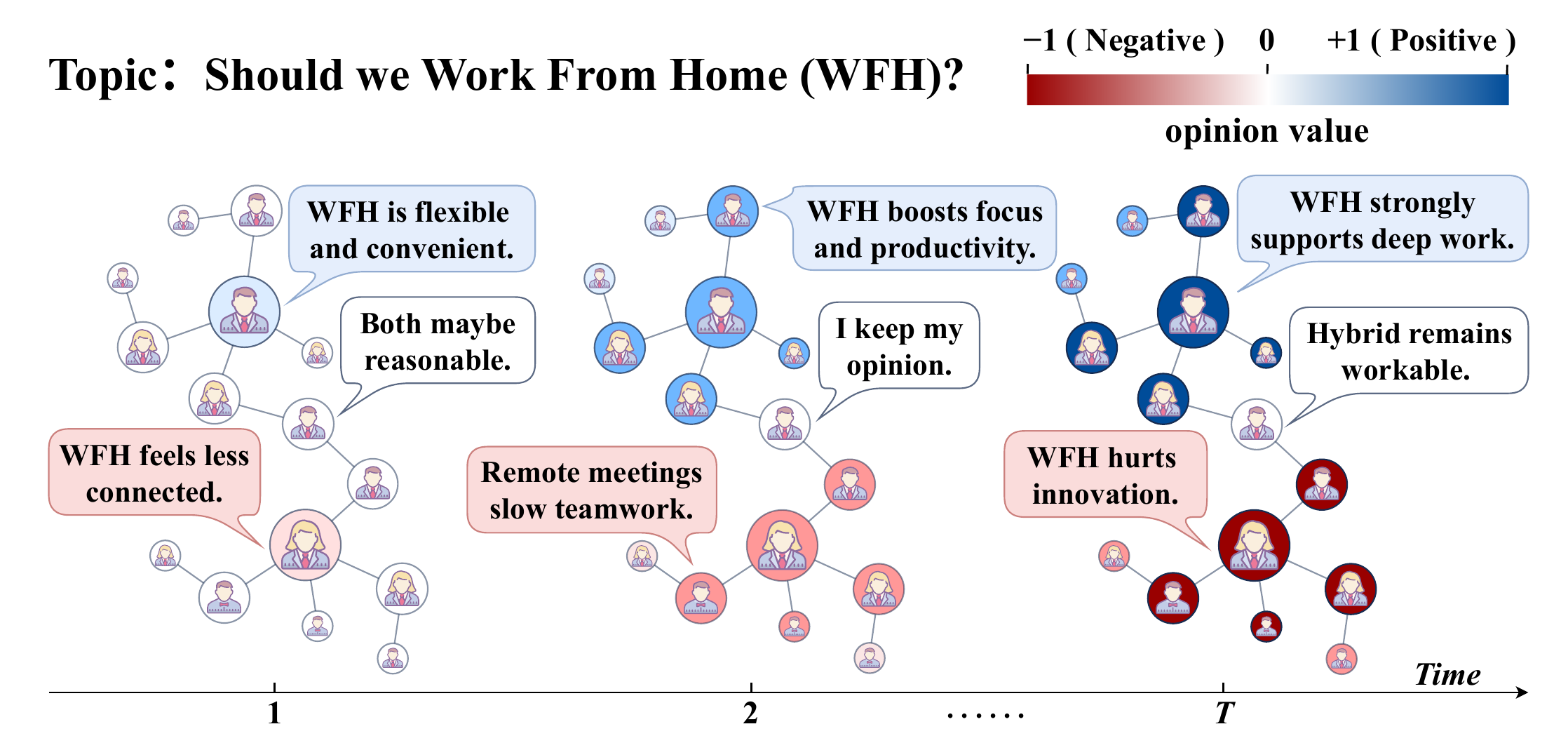} 
    \caption{A conceptual illustration of social opinion evolution. Given an information item, users interact within the social platforms and update their opinions over time.} 
    \label{example} 
\end{figure}

Opinion dynamics centers on unraveling how information dissemination shapes the opinion evolution within human society~\cite{okawa2022predicting}.
Online media has revolutionized the paradigm of information spread, enabling the public to comment on trending events and exchange viewpoints continuously ~\cite{o2010tweets}, thereby driving the evolution of social opinions as shown in Figure~\ref{example}.
Opinion evolution on social platforms is modulated by multiple factors: the topology of social networks~\cite{li2025unigo}, the impacts of social gravity and external interventions~\cite{chandrasekaran2025network}, and users’ intrinsic characteristics~\cite{shirzadi2025opinion}.
Such complexity motivates us to conceptualize opinion evolution as an intricate dynamical system~\cite{yang2025sociologically}.
Understanding this system's underlying mechanisms and predicting its trajectory has become a key issue in social computing, and is crucial across marketing tactics~\cite{ye2012exploring}, public-voice monitoring~\cite{lai2018stance,santos2021link}, and political security~\cite{bond201261,anstead2015social}.
Representative studies into opinion dynamics modeling primarily utilize mechanical and data-driven methods, both of which suffer from inherent limitations.
Mechanical methods specify opinion evolution based on user interactions via difference equations, with classic opinion dynamics including the DeGroot~\cite{degroot1974reaching}, French-Jackson~\cite{friedkin1990social}, and Hegselmann-Krause~\cite{hegselmann2015opinion} models.
Although these models are interpretable in sociology, their oversimplified mechanisms are inadequate for complex and volatile real-world scenarios~\cite{duan2025bi}.
Conversely, data-driven models directly uncover patterns within data from social networks to predict future trends~\cite{qiu2018deepinf,li2025unigo}.
While flexible, the lack of inductive bias introduced by opinion dynamics may result in incomplete or misleading representations~\cite{okawa2022predicting}.
Recently, the advent of Physics-Informed Neural Networks (PINNs) facilitates the integration of flexible neural networks and physical knowledge constraints~\cite{raissi2019physics,karniadakis2021physics}.
Meanwhile, the opinion evolution can be naturally framed as a physical dynamical system for interpretation and analytical purposes~\cite{yang2025sociologically}.
Therefore, a central proposition emerges:
\textbf{Can we advance opinion modeling with PINNs to synergize dynamical interpretability with data-driven flexibility?} 

Despite this promise, the integration of physics priors with opinion modeling remains an open issue, encumbered by the following challenges.
Previous studies introduce classical opinion dynamics to constrain neural networks~\cite{yang2025sociologically,xu2025sigrl}, which can be viewed as a closed-system assumption rooted in limited interaction kinetics~\cite{lv2023unified}.
Such fragmented priors isolate local assimilation, global influence and individual traits, ignoring their intricate interplay in open-system landscape.
The discrepancy between reductive physical priors and multifaceted complexity of social scenarios widens the conceptual gap, yielding an incomplete outline of opinion modeling. 
Therefore, the first challenge lies in: \textbf{How to derive a comprehensive physical system to characterize real-world opinion dynamics?} 
To enforce physical constraints on neural networks, existing methods formulate system dynamics as ordinary differential equations (ODEs) and incorporate them into loss function as penalty terms~\cite{okawa2022predicting,yang2025sociologically}.
This way inevitably suffers from the gradient pathology between data fidelity and physical regularization terms during optimization~\cite{wang2021understanding}.
Meanwhile, introducing external constraints rather than architectural inductive bias decouples latent representations from explicit physical variables, hindering the inherent interpretability of physics-informed models~\cite{karniadakis2021physics}.
To this end, another challenge arises:
\textbf{How to coordinate neural networks with physical priors to unify flexibility and interpretability?}

In this paper, we propose \textsc{Opinn}, a novel physics-informed neural framework for opinion dynamics modeling.
For the first challenge, we draw upon the interacting particle theory~\cite{de1986reaction} and adapt the Diffusion-Convection-Reaction (DCR) system~\cite{codina1998comparison} to interpret opinion modeling through a physical lens.
Specifically, we decompose real-world opinion dynamics into local, global, and endogenous levels, and then map them to the diffusion, convection, and reaction components of the DCR system, respectively.
In addition, we offer a conceptual bridge for this system with classical opinion dynamics model from a theoretical perspective.
For the second challenge, we glean insights from Neural ODEs~\cite{chen2018neural} and formulate neural networks as continuous-time dynamical systems.
In particular, we parameterize the diffusion and convection processes via the graph convolution operator and customized attention mechanism, respectively, while exploring diverse realizations of the reaction term.
Moreover, we introduce learnable coefficients to adaptively modulate the contribution of each branch, ultimately yielding the formulation of neural opinion dynamics.
This paradigm guarantees the compliance with built-in system constraints while ensuring that learned representations preserve established physical semantics.
Building on the above, we present the \textsc{Opinn} framework, comprising an autoencoder architecture integrated with a neural dynamics module.
Our contributions are summarized as follows:

\begin{itemize}
    \item 
    We offer a physical view to interpret opinion modeling, mapping real-world opinion dynamics to each components of the DCR system. 
    Moreover, we provide a conceptual bridge for this system with classical opinion dynamics models.

    \item 
    We implement the DCR system in the form of Neural ODEs, and define the neural opinion dynamics. 
    Grounded in this paradigm, we present the \textsc{Opinn}, a physics-informed neural framework tailored for opinion dynamics modeling.

    \item Through comprehensive evaluations on both synthetic and real-world datasets,  we demonstrate that \textsc{Opinn} consistently outperforms existing state-of-the-art baselines in terms of predictive performance and model generalization.
\end{itemize}

\section{Related Work}
\subsection{Opinion Dynamics Modeling}
\noindent \textbf{Mechanical methods.}
Early studies characterize how individuals update opinions through predefined interaction mechanisms~\cite{sznajd2000opinion}.
The most typical example is the Voter model~\cite{yildiz2010voting}, which assumes that users in the system randomly adopt the opinions of others until convergence.
Based on the local opinion assimilation, the DeGroot model formalizes opinion evolution as a weighted averaging process~\cite{degroot1974reaching} within social networks, while FJ (Friedkin-Johnsen) model further incorporates a stubbornness coefficient to capture scenarios where individuals retain initial beliefs~\cite{friedkin1990social}.
Furthermore, HK (Hegselmann-Krause) model introduces confidence bounds to model global bias assimilation, whereby individuals only interact with those whose opinions fall within a specific range~\cite{hegselmann2015opinion}.
Drawing upon social psychology, later studies explore more advanced update rules involving group effect~\cite{hou2021opinion} and deceptive behavior~\cite{barrio2015dynamics}.
While interpretable, mechanical methods inherently suffer from reductionist and incomplete mechanism, ignoring the intricate interplay of diverse interaction patterns in real-world scenarios~\cite{duan2025bi}. 

\noindent \textbf{Data-driven methods.}
The explosion of social media fuels opinion modeling with data support, driving a shift toward data-driven methods powered by deep learning.
Given the significance of social topology~\cite{de2019learning}, DeepInf~\cite{qiu2018deepinf} reveals that inductive bias introduced by graph neural networks can improve social dynamics prediction~\cite{kipf2016semi,velivckovic2017graph}.
To mine richer structural information, UniGO~\cite{li2025unigo} adopts a graph coarsening-refinement strategy and aggregate super-node messages to uncover patterns beyond neighborhood.
Further, iTransformer~\cite{liu2023itransformer} and SGFormer~\cite{wu2023sgformer} utilize Transformer~\cite{vaswani2017attention} architecture to extend local patterns to global aggregation for long-range dependencies~\cite{gong2026survey,muller2023attending}.
Despite this promise, data-driven paradigm faces inherent limitations: 
due to dependency on data quality, these methods underperform in noisy social platforms without explicit mechanism constraints, potentially introducing erroneous inductive biases.
In addition, the ``black-box'' nature of neural networks hinders the interpretability of opinion evolution analysis~\cite{karniadakis2021physics}.
 
\subsection{Physics-Informed Neural Networks}
To bridge flexibility with interpretability, PINNs have proven effective in integrating scientific priors into neural networks to solve complex problems~\cite{raissi2019physics}.
Modeling opinions from a physical system lens, SINN~\cite{okawa2022predicting} pioneers the reformulation of classic dynamics (DeGroot, FJ, and HK models) to inform neural networks.
ODENet~\cite{lv2023unified} further reveals the underlying relation between opinion dynamics such as confidence bounded models~\cite{hegselmann2015opinion} and neural diffusion~\cite{chamberlain2021grand}.
While grounded in sociology, such priors treat local, global, and endogenous interactions in isolation, ignoring their intricate interplay in real-world scenarios~\cite{choi2023gread}.  
This gap necessitates a comprehensive physical system~\cite{codina1998comparison} to integrate multifaceted interaction patterns for opinion dynamics modeling.
Moreover, existing studies~\cite{okawa2022predicting,yang2025sociologically} recast system priors as an ODE-based regularization to penalize the loss function, which induces the gradient pathology issue~\cite{wang2021understanding} and decouples latent representations from explicit physical meanings~\cite{karniadakis2021physics}. 
Inspired by Neural ODEs~\cite{chen2018neural}, recent advancements formulate neural networks as continuous-time dynamical systems to establish PINNs~\cite{li2024predicting}.
This paradigm guides neural modules to explore within a physically consistent space and endows latent representations with physical semantics, having been widely applied in network science~\cite{chamberlain2021grand,choi2023gread,wu2023advective} and atmospheric physics~\cite{hettige2024airphynet,verma2024climode}.

\section{Preliminaries}
\subsection{Problem Definition}
We define the social network as a graph $\mathcal{G} = (\mathcal{V}, \mathcal{E})$, where $V$ and $E$ are the sets of users and social relations. 
The social topology is described by its adjacency matrix $\mathbf{A}\in \{ 0,1\}^{N\times N}$, where $N = |\mathcal{V}|$ is the number of users.
For a given topic, the user opinion profile at time $t$ is denoted by $\mathbf{X}(t) = [x_1(t),...,x_N(t)]\in\mathbb{R}^{N}$. 
The component $x_i(t)\in[-1,1]$ quantifies the stance of user $i$ on a continuous scale from -1 (negative) to 1 (positive).
We formulate opinion dynamics modeling as a time-series forecasting problem within social network.
Given the sequence $\mathbf{X}_{[1 \sim T]} = \{\mathbf{X}(1),..., \mathbf{X}(T)\} \in \mathbb{R}^{N\times T}$ during time-window $[1,T]$, the objective is to learn a function $\mathcal{F}(\cdot)$ that can predict future opinion trends over the horizon $h$:
\begin{equation}
    \hat{\mathbf{X}}_{[T+1 \sim T+h]} = \mathcal{F}(\mathbf{X}_{[1 \sim T]},\mathbf{A}) \in \mathbb{R}^{N \times h},
\end{equation}
where $\hat{\mathbf{X}}_{[T+1 \sim T+h]}$ represents the prediction of user opinions.  

\subsection{Diffusion-Convection-Reaction System}
Derived from the continuity equation, the DCR equation~\cite{codina1998comparison} captures the mass conservation within interacting particle systems:
\begin{equation}
    \frac{\partial z}{\partial t} = \underbrace{\kappa \nabla^2 z}_{Diffusion} - \underbrace{\vec{v} \cdot \nabla z}_{Convection} + \underbrace{r(z,t)}_{Reaction},
\end{equation}
where $\frac{\partial z}{\partial t}$ is change rate of concentration $z$ to time $t$.
The diffusion term, characterized by the diffusion coefficient $\kappa$ and Laplacian operator $\nabla^2$, quantifies the local movement of particles down the concentration gradient, as defined by Fick's Law~\cite{bird2002transport}.
Convection, expressed via the velocity vector $\vec{v}$ and concentration gradient $\nabla z$, describes the bulk transport of particles driven by global velocity field~\cite{bird2002transport}.
Unlike spatial transport described by diffusion and convection, the reaction term $r(z,t)$ accounts for endogenous dynamics.
It focuses on the internal growth or decay of concentration, including source and sink processes that reflect intrinsic state changes~\cite{choi2023gread}.

\subsection{Neural Ordinary Differential Equations}
Neural ODEs adopt neural networks to fit the vector field of ordinary differential equations~\cite{chen2018neural}, offering a flexible way to model the trajectory of dynamical system.
Given the system with an initial state $z(0)$, the state $z(T)$ at time $T$ can be obtained by integrating the system's dynamical equation over the interval $[0,T]$:
\begin{equation}
    z(T)- z(0) = \int^{T}_{0} \frac{dz(t)}{dt} \, dt = \int^{T}_{0} f(z(t), t, \theta) \, dt, 
\end{equation}
where $f(\cdot)$ is a neural function parameterized by $\theta$ to approximate the vector field $\frac{dz(t)}{dt}$.
This paradigm embeds the architectural inductive bias to enforce system constraints on the neural network, while ensuring that the learned representations retain established physical semantics~\cite{bilovs2021neural}.
Established ODE solvers such as Runge-Kutta or Dormand–Prince methods~\cite{dormand1980family} can be used to compute this integral, and the forward pass is formulated as:
\begin{equation}
    z(t) = \text{ODESolver}(z(0), [0,...,t], f, \theta).
\end{equation}

\begin{figure}[htbp]
    \centering 
        \includegraphics[width=0.48\textwidth]{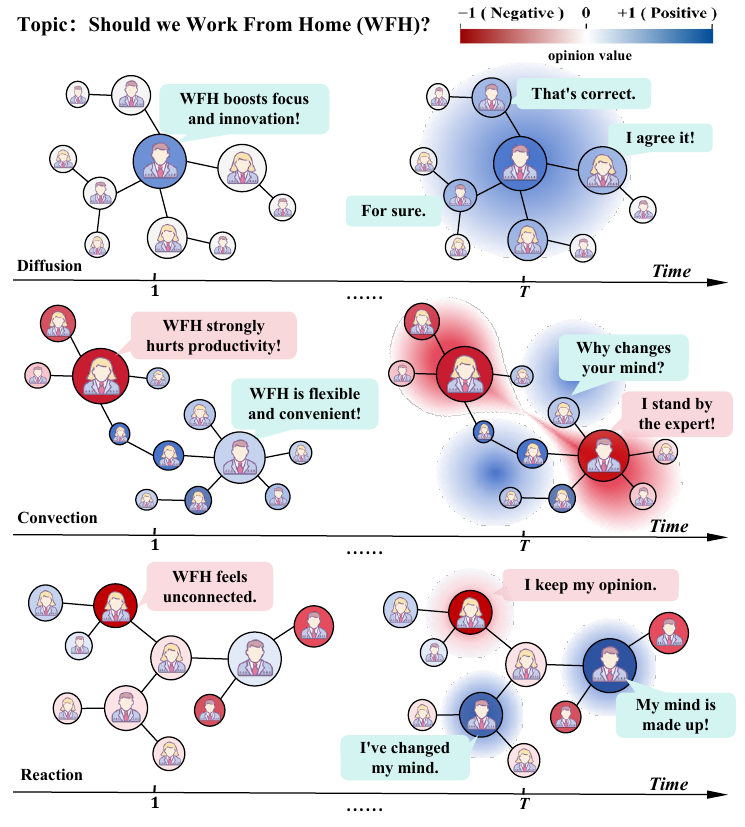} 
    \caption{The conceptual schematic of introducing DCR system into opinion dynamics modeling: diffusion describes the local consensus of user opinions, convection captures the drift driven by external force field, while reaction depicts the opinion evolution dictated by user internal drivers.} 
    \label{mov} 
\end{figure}
\begin{figure*}[ht]
    \centering 
        \includegraphics[width=0.82\textwidth]{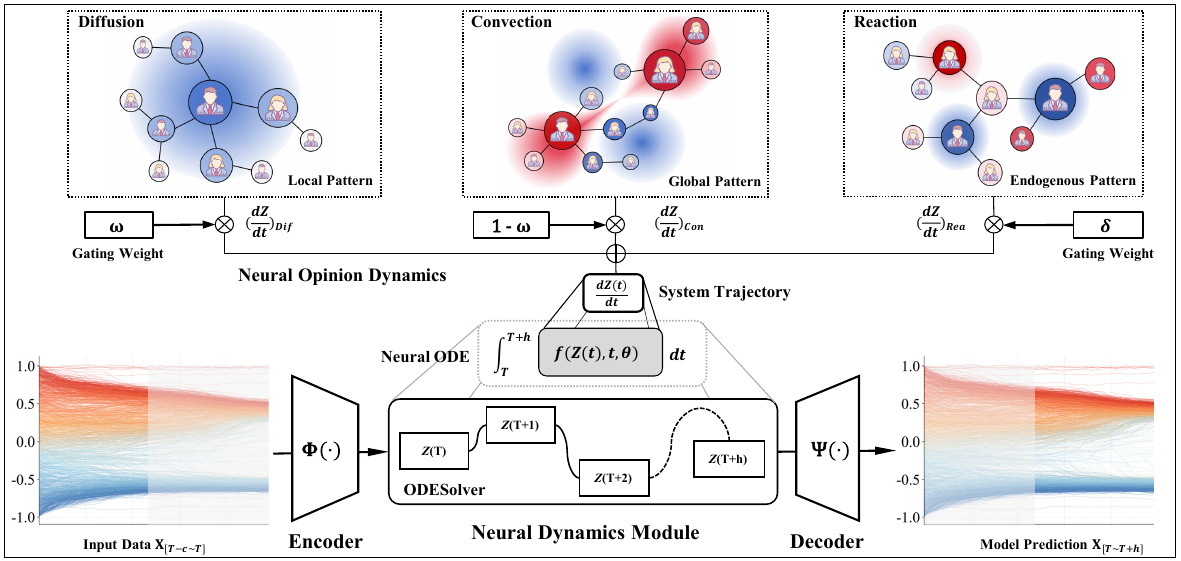} 
    \caption{The overall architecture of \textsc{Opinn} comprises three core components: (1) an encoder to map the observed user opinions into latent system states; (2) a neural dynamics module to model the trajectory of opinion dynamical system; and (3) a decoder to generate opinion predictions by integrating neural representations with dynamical priors.} 
    \label{framework} 
\end{figure*}

\section{Methodology}

\subsection{Physics-Informed Opinion Modeling}
From a physical perspective, opinion evolution is inherently a complex system constituted by an ensemble of interacting particles (users)~\cite{yang2025sociologically}.
Classical opinion dynamics typically compartmentalize local, global, and endogenous interactions, neglecting the intricate interplay among these three levels (see Appendix~\ref{appa1} for theoretical basis).
To bridge this gap, we view the state of opinion evolution as the flux~\cite{verma2024climode} inspired by interacting particle theory~\cite{de1986reaction}, and introduce the DCR system formulated by ODE:
\begin{equation}
    \left(\frac{d x}{d t}\right)_{DCR} = \left( \frac{d x}{d t} \right)_{Dif} + \left(\frac{d x}{d t}\right)_{Con} + \left(\frac{d x}{d t}\right)_{Rea},
\end{equation}
which provides a comprehensive system across three distinct scales to accommodate real-world opinion dynamics.
Next, we detail the mapping between DCR components and practical scenarios as illustrated by the conceptual schematic in Figure~\ref{mov}, while connecting this dynamical system with established opinion dynamics models.

\noindent \textbf{Diffusion.}
Analogous to the gas or heat diffusion in physics, the diffusion of opinions reveals the pattern of gradual consensus within social networks~\cite{chamberlain2021grand}.
By employing the spatial discretization~\cite{bronstein2017geometric}, we can obtain the ODE-based version of the diffusion process:
\begin{equation}
\label{dif}
    \left( \frac{d x_i}{d t} \right)_{Dif} =  \sum_{j \in \mathcal{N}(i)} w_{ij}( x_j(t)- x_i(t)),
\end{equation}
where $x_i(t)$ denotes the opinion of user $i$ at time $t$, $\mathcal{N}(i)$ denotes its neighboring set, and $w_{ij}$ denotes the transport velocity from user $i$ to $j$.
Then, we transform this equation into a recurrence version to explain its connection with classical opinion models:
\begin{equation}
    x_i(t+1) = (1 - \sum_{j \in \mathcal{N}(i)} w_{ij}) x_i(t) + \sum_{j \in \mathcal{N}(i)} w_{ij} x_j(t),
\end{equation}
where the first term modulates the user's internal state, while the second integrates the neighborhood information.
Therefore, diffusion characterizes the local assimilation pattern as a generalization of the DeGroot model~\cite{degroot1974reaching}, with formal derivations and analysis of this conceptual connection detailed in Appendix~\ref{appa2}.

\noindent \textbf{Convection.}
The opinion convection refers to the global and directional propagation driven by social gravity or external mechanisms~\cite{wu2023advective}, similar to the pollution migration caused by river currents or atmospheric flow~\cite{hettige2024airphynet}.
Referring to~\cite{chapman2015advection}, the ODE-based version of convection can be expressed as:
\begin{equation}
   \label{con}
   \left(\frac{d x_i}{d t}\right)_{Con} = \sum_{\forall j | j\rightarrow i} \vec{w}_{ji} x_j(t)  - \sum_{\forall k | i \rightarrow k} \vec{w}_{ik} x_i(t), 
\end{equation}
where $\vec{w}_{ji}$ represents the directed velocity of convection from user $j$ to $i$ ($j\rightarrow i, j \neq i$).
Conceptually, convection formalizes the directional drift of opinions, accounting for updates driven by exogenous fields or macro-level steering, such as policy intervention~\cite{lai2018stance} and group effect~\cite{hou2021opinion}.
Its iterative scheme can be formulated as:
\begin{equation}
    x_{i}(t+1) = (1- \sum_{k \in \mathcal{V} \setminus i}\vec{w}_{ik}) \  x_i(t) + \sum_{j \in  \mathcal{V} \setminus i}\vec{w}_{ji} \ x_j(t).
\end{equation}
By modulating the flow velocity for interaction filtering, the HK model~\cite{hegselmann2015opinion} can be interpreted as a selective convection constrained by confidence bound, as formally proven in Appendix~\ref{appa2}.

\noindent \textbf{Reaction.}
Distinct from the spatial transport processes of diffusion and convection, reaction captures the endogenous shifts within individuals~\cite{choi2023gread}, analogous to the internal chemical processes such as material growth, decay, and combustion:
\begin{equation}
    \left(\frac{d x_i}{d t}\right)_{Rea} = r(x_i,t)
\end{equation}
where $r(\cdot)$ is the reaction term.
The selection of this term for opinion modeling is flexible: it can be formulated as the initial opinion $x_i(0)$ to represent belief persistence, paralleling the stubbornness mechanism in FJ model~\cite{friedkin1990social} (see derivations in Appendix~\ref{appa2}).
Moreover, the reaction term can scale linearly with $x_i(t)$ to describe anchoring or relaxation dynamics, while exploration of non-linear reaction remains a compelling direction for future research~\cite{choi2023gread}.

\subsection{Neural Opinion Dynamics}
\label{neuralode}
Building upon the DCR system, we aim to synergize the physics-inspired interpretability with data-driven flexibility to implement it.
Inspired by Neural ODEs~\cite{chen2018neural}, we employ neural networks to parameterize the DCR components, thereby grounding the neural representations in physical principles~\cite{bilovs2021neural}.
Furthermore, we define the neural opinion dynamics by coupling three processes via gating weights to steer the modeling of opinion evolution.

\noindent \textbf{Diffusion.}
Inspired by neural diffusion on graphs~\cite{chamberlain2021grand}, we view the social topology as the deterministic diffusivity and parameterize diffusion via the graph convolutional operator~\cite{kipf2016semi}.
Given $\mathbf{Z}(t)$ as the system state comprising all user's opinions at time $t$, the diffusion term is defined as:
\begin{equation}
    \left(\frac{d \mathbf{Z}(t)}{d t}\right)_{Dif} \coloneqq \sigma((\textbf{I} - \tilde{\textbf{L}})\textbf{Z}(t)\textbf{W}_D),
\end{equation}
where $\sigma(\cdot)$ is the activation function, $\textbf{I}$ represents the identity matrix, $\textbf{W}_D$ represents the learnable weight matrix, and $\tilde{\textbf{L}}$ is the symmetric normalized graph Laplacian matrix derived from $\textbf{A}$.

\noindent \textbf{Convection.}
Distinct from local diffusion, we extend the social circle to a fully connected graph to realize global convection:
\begin{equation}
    \left(\frac{d \mathbf{Z}(t)}{d t}\right)_{Con} \coloneqq \text{softmax}(\vec{\textbf{V}}) \textbf{Z}(t)\textbf{W}_C,
\end{equation}
where $\vec{\textbf{V}}$ denotes the velocity matrix.
To model opinion-dependent update intensity, we parameterize the directed velocity between users via a customized attention mechanism:
\begin{equation}
    \vec{\textbf{V}}_{ij} = \sigma((\textbf{z}_i(t)- \textbf{z}_j(t))\textbf{W}_V),
\end{equation}
where $\textbf{z}_i(t)$ represent the state vector for user $i$. 
The velocity is governed by disparities in opinion states inspired by confidence bound models~\cite{hegselmann2015opinion}, capturing the directional driving force of opinion drift.

\noindent \textbf{Reaction.}
Given the endogenous intricacies of real-world users, we consider the following schemes to realize the reaction:
\begin{equation}
    \left(\frac{d \textbf{Z}(t)}{d t}\right)_{Rea} \coloneqq 
    \begin{dcases*}
    \ \textbf{Z}(t) & \text{if Source term,}  \\
    \ \text{Linear}(\textbf{Z}(t)) & \text{if Linear term,} \\
    \ \text{MLP}(\textbf{Z}(t)) & \text{if Non-linear term,} \\
\end{dcases*}
\end{equation}
where $\textbf{Z}(t)$ is the current system state, $\text{Linear}(\cdot)$ denotes a single linear layer, and $\text{MLP}(\cdot)$ denotes a multi-layer perceptron.
Established mechanisms such as Allen-Cahn~\cite{allen1979microscopic} process are applicable for polarization dynamics~\cite{wang2022acmp}.
Here, we summarize these as source, linear, and non-linear term within our scope (see experiments in Section~\ref{ma}), and leave deeper explorations to future works~\cite{choi2023gread}.

\noindent \textbf{Neural opinion dynamics.}
Given the intricate interplay among various opinion dynamics in real-world scenarios, we further introduce learnable coefficients $\omega$ and $\delta$ to modulate each branch within DCR system, yielding the formulation neural opinion dynamics:
\begin{equation}
    \frac{d \textbf{Z}}{d t} = \omega \cdot \left( \frac{d \textbf{Z}}{d t} \right)_{Dif} + (1-\omega) \cdot\left(\frac{d \textbf{Z}}{d t}\right)_{Con} + \delta \cdot \left(\frac{d \textbf{Z}}{d t}\right)_{Rea}.
\end{equation}
To capture the complex coupling of local and global dynamics, the fusion of diffusion and convection transport is governed by a gating weight $\omega \in [0,1]$.
Additionally, the user's endogenous pattern is prioritized through a separate gating weight $\delta$.

\subsection{The \textsc{Opinn} Framework}
Grounded in neural opinion dynamics, we further present \textsc{Opinn}, a
physics-informed neural framework for predicting opinion trends.
As shown in Figure~\ref{framework}, the architecture comprises three components: an encoder, a neural dynamics module, and a decoder.

\noindent \textbf{Encoder.}
Given the noisy and redundant input space, we follow~\cite{chamberlain2021grand,lv2023unified,hettige2024airphynet} by mapping user opinions into latent space for dynamical modeling.
Specifically, we employ a neural module $\Phi(\cdot)$ to encode the historical opinions for all users as the system state:
\begin{equation}
    \textbf{Z}(t) = \Phi(\textbf{X}_{[t-c\sim t]}) \in \mathbb{R}^{N\times D},
\end{equation}
where $\textbf{Z}(t)$ approximates the system state~\cite{verma2024climode,hettige2024airphynet} at time $t$, $c$ denotes the length of input context, and $D$ is the hidden dimension.
To capture the temporal dependencies of opinion dynamics, we implement the encoder by a Gated Recurrent Unit (GRU) module~\cite{chung2014empirical} within \textsc{Opinn} (see comparative experiments in Section \ref{ablationex}).

\noindent \textbf{Neural dynamics module.}
Based on the proposed neural opinion dynamics, we introduce a Neural ODE module to characterize the evolution of opinion dynamical system.
Given the system state $\textbf{Z}(t)$, the vector field is parameterized via neural functions:
\begin{equation}
    \frac{d \textbf{Z}(t)}{dt} := f(\textbf{Z}(t),t,\theta),
\end{equation}
where $f(\cdot)$ is $\theta$-parameterized to approximate the system trajectory, consistent with the neural opinion dynamics in Section~\ref{neuralode}.
Based on the dynamical trajectory and system state $\textbf{Z}(T)$ at time $T$, we adopt the off-the-shelf ODE solvers to obtain future states:
\begin{equation}
    \hat{\textbf{Z}}_{[T+1 \sim T+h]} = \text{ODESolver}(\textbf{Z}(T), [T,...,T+h], f, \theta) \in \mathbb{R}^{N \times D \times h}.
\end{equation}

\noindent \textbf{Decoder.}
Subsequently, the future system states are mapped back to the original space via a decoder to generate predictions:
\begin{equation}
    \hat{\textbf{X}}_{[T+1\sim T+h]} = \Psi (\hat{\textbf{Z}}_{[T+1\sim T+h]}) \in \mathbb{R}^{N \times h},
\end{equation}
where $\Psi(\cdot)$ is implemented by a two-layer MLP module.
Finally, the model is trained using the Mean Squared Error (MSE) loss as the optimization criterion.

\begin{figure*}[htbp]
    \centering
    \begin{subfigure}[b]{0.3\linewidth}
        \centering
        \includegraphics[width=\linewidth]{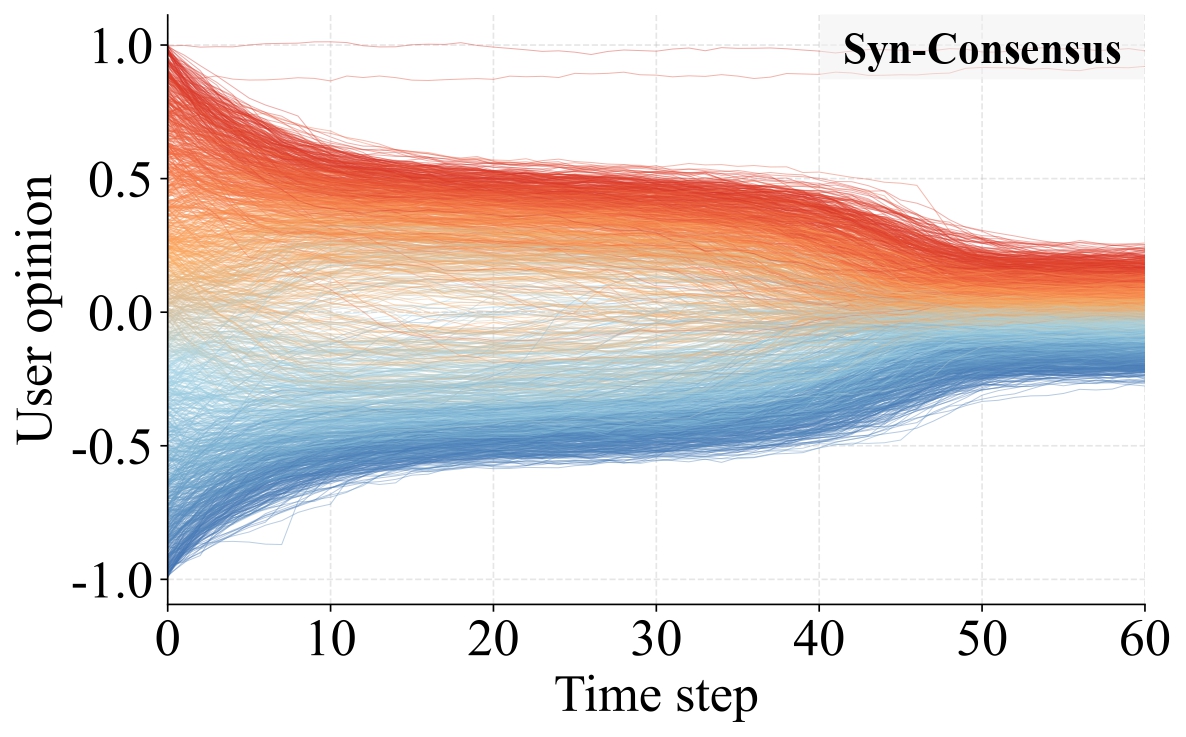}
    \end{subfigure}
    \hfill
    \begin{subfigure}[b]{0.3\linewidth}
        \centering
        \includegraphics[width=\linewidth]{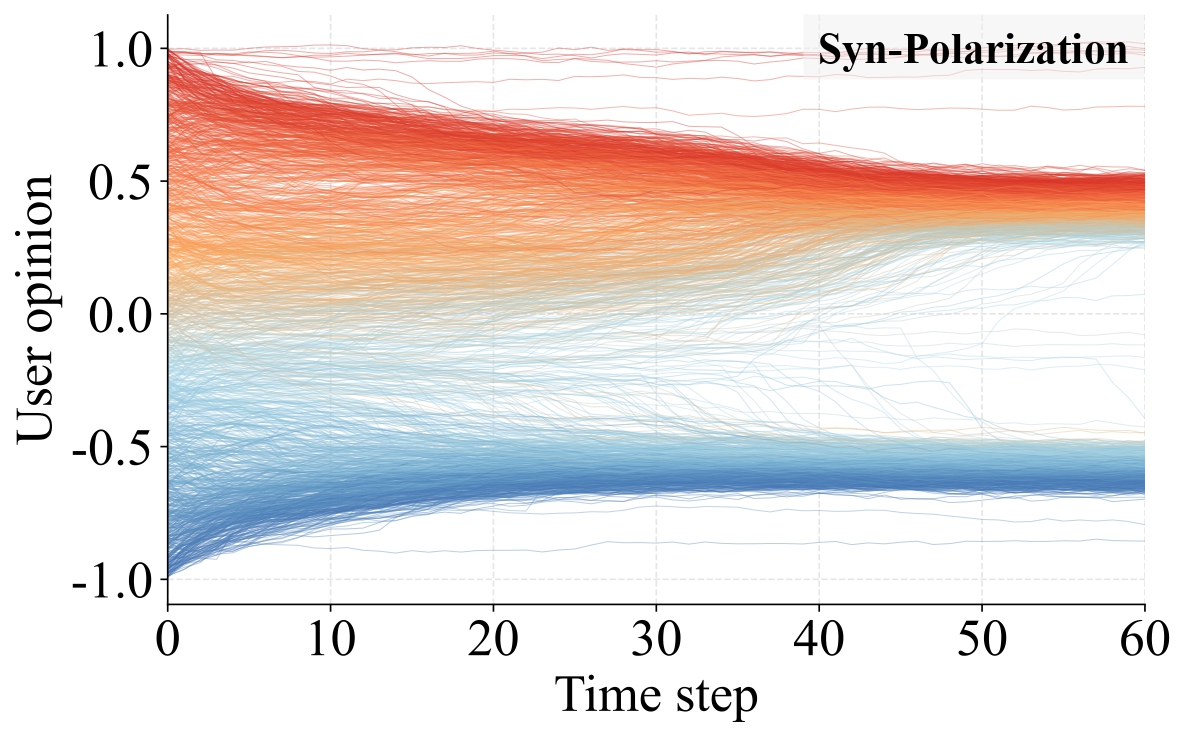}
    \end{subfigure}
    \hfill
    \begin{subfigure}[b]{0.3\linewidth}
        \centering
        \includegraphics[width=\linewidth]{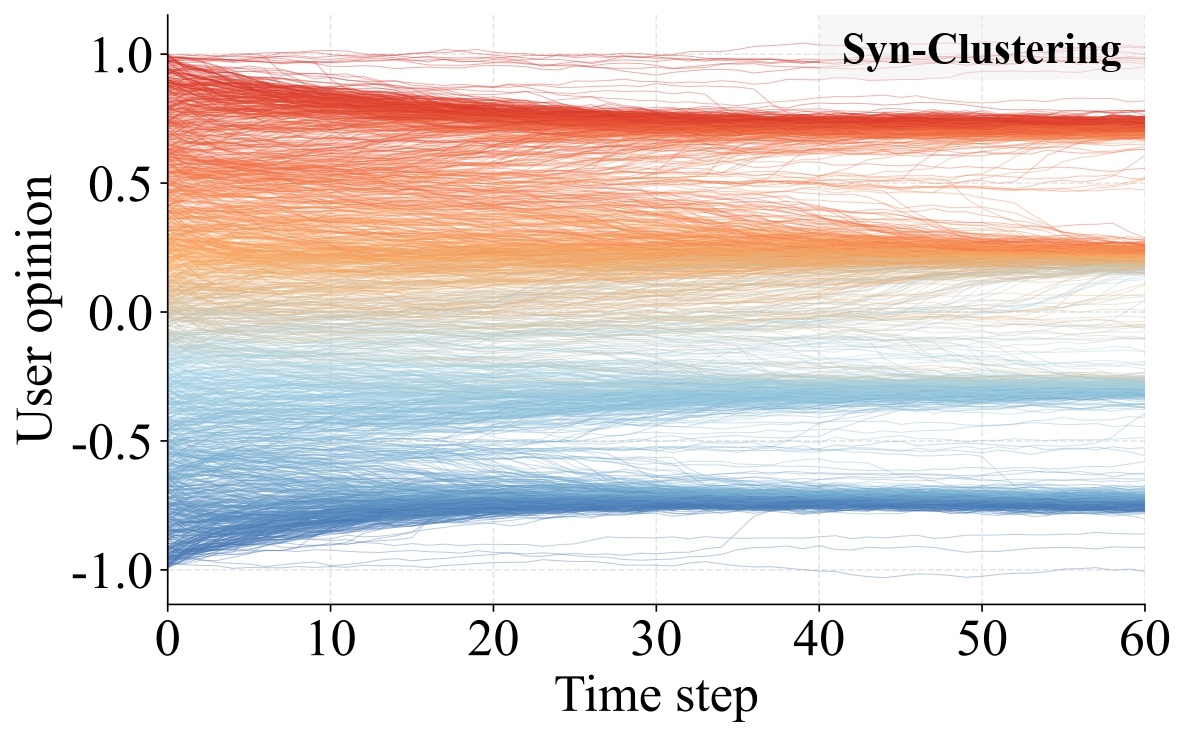}
    \end{subfigure}
    \caption{Visualization of synthetic datasets: we randomly sampled 1,500 users to demonstrate the overall evolutionary trends.} 
    \label{syndata}
\end{figure*}

\begin{table}[htbp]
    \setlength{\tabcolsep}{4pt}
    \centering
    \caption{The basic statistics of four real-world datasets. }
    \label{realdatasets}
    \resizebox{0.45\textwidth}{!}{
    \begin{tabular}{c c c c c c}
    \toprule
    \textbf{Dataset} & \textbf{\# Nodes} & \textbf{\# Edges} & \textbf{\# Posts} & \textbf{Time span} & \textbf{Time steps}\\
    \midrule
    Delhi Election & 548 & 5,271 & 20,026 & 2013/12/9 - 2013/12/15 & 547 \\
    U.S. Election & 526 & 2,482 & 10,883 & 2023/12/1 -  2024/3/5 & 393 \\
    Israel-Palestine & 10,695 & 89,235 & 201,365 & 2023/10/7 - 2023/10/19 & 265 \\
    COVID-19 & 894 & 21,143 & 102,635 & 2020/1/1 - 2020/3/31 & 386 \\
    \bottomrule
    \end{tabular}
    }
\end{table}

\section{Experiments}
\subsection{Data Description}
To comprehensively evaluate \textsc{Opinn}, we conduct experiments using both real-world datasets and synthetic datasets.

\noindent \textbf{Real-world datasets.}
We use four real-world datasets released in~\cite{li2025unigo,de2019learning}, including Delhi Election, U.S. Election, Israel-Palestine, and COVID-19.
The COVID-19 is collected from Weibo, while the others are sourced from X (Twitter).
Irrelevant accounts and social bots are filtered out, and only posts from active users regarding the topic are selected.
Opinion values are jointly labeled manually and with Large Language Models (LLMs), using a continuous scale from [-1,1] to indicate the user's stance (from negative to positive) based on posts to specific topics.
The social network is static and constructed based on user interactions, specifically the retweets and comments.
Basic statistics of real-world datasets is presented in Table~\ref{realdatasets}, with further details presented in Appendix~\ref{appdata1}.

\noindent \textbf{Synthetic datasets.}  
Complementary to the four real-world datasets, we synthesize opinion datasets under three typical patterns: consensus, polarization, and clustering.
The simulation involves 10,000 users whose initial opinions are uniformly distributed~\cite{okawa2022predicting}, and social structure is built upon a BA scale-free graph~\cite{barabasi1999emergence} with about 100,000 edges.
We modulate the opinion dynamics by referring to the update rule in~\cite{li2025unigo}, and visualization is shown in Figure~\ref{syndata}.
We perform linear interpolation on the non-converging phase and set the time step to 400.
More details about the synthetic rules and implementation can be found in Appendix~\ref{appdata2}.

\subsection{Experimental Settings}

\noindent \textbf{Baselines.}
We compare \textsc{Opinn} with 16 baselines from three categories: 
[\textit{Mechanical methods}]: Voter~\cite{yildiz2010voting}, DeGroot~\cite{degroot1974reaching}, FJ (Friedkin–Johnsen model)~\cite{friedkin1990social}, HK (Hegselmann-Krause model)~\cite{hegselmann2015opinion}.
[\textit{Data-driven methods}]: GCN~\cite{kipf2016semi}, GAT~\cite{velivckovic2017graph}, DeepInf~\cite{qiu2018deepinf}, iTransformer~\cite{liu2023itransformer}, SGFormer~\cite{wu2023sgformer}, UniGO~\cite{li2025unigo}.
[\textit{Physics-informed methods}]: SINN~\cite{okawa2022predicting}, SIGNN~\cite{yang2025sociologically}, ODENet~\cite{lv2023unified}, GRAND~\cite{chamberlain2021grand}, GREAD~\cite{choi2023gread}, AdvDifformer~\cite{wu2023advective}.
Baseline details are presented in Appendix~\ref{appbaseline1}, while a physical perspective analysis is provided in Appendix~\ref{appbaseline2}.

\noindent \textbf{Settings \& Evaluation.}
For real-world datasets, we conduct predictions across two horizons: 30 and 60 time steps (abbreviated as 30\ T and 60\ T), while the horizon for synthetic datasets is fixed at 30 T.
By default, we split datasets chronologically in the ratio of 60\%: 20\%: 20\% to generate training, validation, and test sets.
We employ Mean Absolute Error (MAE) and Root Mean Square Error (RMSE) as evaluation metrics  and run each deep learning model five times to report the average results.

\noindent \textbf{Implementation.}
We implement \textsc{Opinn} using PyTorch 2.4.0 on NVIDIA GeForce RTX PRO 6000 GPU with Adam optimizer.
We employ a GRU module as the encoder and a two-layer MLP as the decoder.
For neural dynamics module, we implement diffusion via a single GCN layer~\cite{kipf2016semi} and convection through the described attention layer, while selecting the best-performing reaction term to report the results.
We define the input context as 30 T and treat it as one time step for the system state. 
For ODE solver, we employ the fourth-order Runge-Kutta (RK-4) method with an integration interval of $t=1.0$.
Hyperparameters including the encoder hidden dimension, batch size, learning rate are determined via a grid search.
For more implementation details, please refer to the Appendix~\ref{appimp}.

\begin{table*}[ht]
    \caption{The overall performance comparison of opinion dynamics modeling on metrics RMSE and MAE ($\times 10^{-2}$). Bold and underlined font show the best and the second best result, respectively.}
    \label{rearesults}
    \small
    \resizebox{\textwidth}{!}{
    \setlength{\tabcolsep}{1mm} 
    \begin{tabular}{c | c c | c c | c c | c c | c c | c c | c c | c c }
    \toprule
    \multirow{4}{*}{\textbf{Model}} & 
    \multicolumn{4}{c|}{\textbf{Delhi Election}} &
    \multicolumn{4}{c|}{\textbf{U.S. Election}} &
    \multicolumn{4}{c|}{\textbf{Israel-Palestine}} &
    \multicolumn{4}{c}{\textbf{COVID-19}} \\ 
    \cmidrule(r){2-17}
    & \multicolumn{2}{c|}{30 T} & \multicolumn{2}{c|}{60 T} &
    \multicolumn{2}{c|}{30 T} & \multicolumn{2}{c|}{60 T} &
    \multicolumn{2}{c|}{30 T} & \multicolumn{2}{c|}{60 T} & 
    \multicolumn{2}{c|}{30 T} & \multicolumn{2}{c}{60 T} \\
    \cmidrule(r){2-17}
    & RMSE & MAE & RMSE & MAE & RMSE & MAE & RMSE & MAE & RMSE & MAE & RMSE & MAE & RMSE & MAE & RMSE & MAE \\
    \midrule
        Voter~\cite{yildiz2010voting} & 21.99\textsubscript{$\pm$0.00} & 16.63\textsubscript{$\pm$0.00} & 22.43\textsubscript{$\pm$0.00} & 17.07\textsubscript{$\pm$0.00} & 98.30\textsubscript{$\pm$0.00} & 65.47\textsubscript{$\pm$0.00} & 99.25\textsubscript{$\pm$0.00} & 66.87\textsubscript{$\pm$0.00} & 38.98\textsubscript{$\pm$0.00} & 29.01\textsubscript{$\pm$0.00} & 52.32\textsubscript{$\pm$0.00} & 39.44\textsubscript{$\pm$0.00} & 59.89\textsubscript{$\pm$0.00} & 43.52\textsubscript{$\pm$0.00} & 60.74\textsubscript{$\pm$0.00} & 44.45\textsubscript{$\pm$0.00} \\
        DeGroot~\cite{degroot1974reaching} & 14.76\textsubscript{$\pm$0.00} & 10.95\textsubscript{$\pm$0.00} & 15.06\textsubscript{$\pm$0.00} & 11.24\textsubscript{$\pm$0.00} & 80.43\textsubscript{$\pm$0.00} & 69.65\textsubscript{$\pm$0.00} & 81.49\textsubscript{$\pm$0.00} & 70.74\textsubscript{$\pm$0.00} & 57.57\textsubscript{$\pm$0.00} & 45.70\textsubscript{$\pm$0.00} & 58.32\textsubscript{$\pm$0.00} & 46.33\textsubscript{$\pm$0.00} & 36.57\textsubscript{$\pm$0.00} & 26.95\textsubscript{$\pm$0.00} & 36.11\textsubscript{$\pm$0.00} & 26.29\textsubscript{$\pm$0.00} \\
        FJ~\cite{friedkin1990social} & 13.62\textsubscript{$\pm$0.00} & 
        9.87\textsubscript{$\pm$0.00} & 14.96\textsubscript{$\pm$0.00} & 10.95\textsubscript{$\pm$0.00} & 58.13\textsubscript{$\pm$0.00} & 49.84\textsubscript{$\pm$0.00} & 59.52\textsubscript{$\pm$0.00} & 50.89\textsubscript{$\pm$0.00} & 40.72\textsubscript{$\pm$0.00} & 31.99\textsubscript{$\pm$0.00} & 44.92\textsubscript{$\pm$0.00} & 35.84\textsubscript{$\pm$0.00} & 28.51\textsubscript{$\pm$0.00} & 21.78\textsubscript{$\pm$0.00} & 28.83\textsubscript{$\pm$0.00} & 21.94\textsubscript{$\pm$0.00} \\ 
        HK~\cite{hegselmann2015opinion} & 15.16\textsubscript{$\pm$0.00} & 11.27\textsubscript{$\pm$0.00} & 15.29\textsubscript{$\pm$0.00} & 11.41\textsubscript{$\pm$0.00} & 54.96\textsubscript{$\pm$0.00} & 39.14\textsubscript{$\pm$0.00} & 59.61\textsubscript{$\pm$0.00} & 41.06\textsubscript{$\pm$0.00} & 50.40\textsubscript{$\pm$0.00} & 36.83\textsubscript{$\pm$0.00} & 53.81\textsubscript{$\pm$0.00} & 41.67\textsubscript{$\pm$0.00} & 27.07\textsubscript{$\pm$0.00} & 21.09\textsubscript{$\pm$0.00} & 26.90\textsubscript{$\pm$0.00} & 20.70\textsubscript{$\pm$0.00} \\ 
    \midrule
        GCN~\cite{kipf2016semi} & 14.60\textsubscript{$\pm$0.13} & 10.90\textsubscript{$\pm$0.09} & 14.98\textsubscript{$\pm$0.09} & 11.18\textsubscript{$\pm$0.10} & 56.39\textsubscript{$\pm$0.18} & 44.90\textsubscript{$\pm$0.29} & 60.10\textsubscript{$\pm$0.11} & 47.78\textsubscript{$\pm$0.63} & 34.49\textsubscript{$\pm$0.24} & 26.21\textsubscript{$\pm$0.27} & 50.12\textsubscript{$\pm$0.27} & 40.03\textsubscript{$\pm$0.34} & 25.04\textsubscript{$\pm$0.07} & 20.28\textsubscript{$\pm$0.08} & 29.81\textsubscript{$\pm$0.08} & 22.93\textsubscript{$\pm$0.11}\\
        GAT~\cite{velivckovic2017graph} & 14.63\textsubscript{$\pm$0.11} & 10.91\textsubscript{$\pm$0.10} & 15.02\textsubscript{$\pm$0.10} & 11.25\textsubscript{$\pm$0.13} & 52.16\textsubscript{$\pm$0.11} & 38.76\textsubscript{$\pm$0.27} & 53.22\textsubscript{$\pm$0.07} & 40.18\textsubscript{$\pm$0.35} & 31.09\textsubscript{$\pm$0.17} & 22.24\textsubscript{$\pm$0.20} & 46.42\textsubscript{$\pm$0.19} & 36.84\textsubscript{$\pm$0.25} & 22.71\textsubscript{$\pm$0.09} & 17.65\textsubscript{$\pm$0.11} & 23.27\textsubscript{$\pm$0.09} & 17.18\textsubscript{$\pm$0.12}\\
        DeepInf~\cite{qiu2018deepinf} & 14.27\textsubscript{$\pm$0.06} & 10.54\textsubscript{$\pm$0.07} & 14.75\textsubscript{$\pm$0.06} & 10.93\textsubscript{$\pm$0.08} & 54.57\textsubscript{$\pm$0.14} & 40.95\textsubscript{$\pm$0.24} & 57.46\textsubscript{$\pm$0.05} & 44.21\textsubscript{$\pm$0.39} & 32.56\textsubscript{$\pm$0.14} & 24.19\textsubscript{$\pm$0.16} & 47.48\textsubscript{$\pm$0.17} & 37.43\textsubscript{$\pm$0.22} & 26.26\textsubscript{$\pm$0.05} & 21.06\textsubscript{$\pm$0.07} & 30.03\textsubscript{$\pm$0.06} & 23.04\textsubscript{$\pm$0.10}\\
        UniGO~\cite{li2025unigo} & 12.08\textsubscript{$\pm$0.04} & 8.15\textsubscript{$\pm$0.08} & 13.91\textsubscript{$\pm$0.06} & 10.15\textsubscript{$\pm$0.10} & \underline{45.39}\textsubscript{$\pm$0.08} & \underline{29.75}\textsubscript{$\pm$0.22} & 50.30\textsubscript{$\pm$0.15} & 36.61\textsubscript{$\pm$0.44} & 30.62\textsubscript{$\pm$0.21} & 22.19\textsubscript{$\pm$0.22} & 42.51\textsubscript{$\pm$0.23} & 33.01\textsubscript{$\pm$0.28} & 18.83\textsubscript{$\pm$0.06} & 14.07\textsubscript{$\pm$0.06} & \underline{20.21}\textsubscript{$\pm$0.07} & \underline{14.83}\textsubscript{$\pm$0.09}\\
        SGFormer~\cite{wu2023sgformer} & 13.35\textsubscript{$\pm$0.04} & 9.70\textsubscript{$\pm$0.05} & 14.43\textsubscript{$\pm$0.03} & 10.60\textsubscript{$\pm$0.05} & 50.14\textsubscript{$\pm$0.06} & 35.66\textsubscript{$\pm$0.15} & 51.81\textsubscript{$\pm$0.04} & 37.79\textsubscript{$\pm$0.33} & 31.56\textsubscript{$\pm$0.10} & 23.09\textsubscript{$\pm$0.12} & 46.23\textsubscript{$\pm$0.17} & 35.98\textsubscript{$\pm$0.19} & 19.78\textsubscript{$\pm$0.04} & 13.59\textsubscript{$\pm$0.04} & 22.65\textsubscript{$\pm$0.05} & 16.77\textsubscript{$\pm$0.05} \\
        iTransformer~\cite{liu2023itransformer} & 12.16\textsubscript{$\pm$0.03} & 8.23\textsubscript{$\pm$0.06} & 13.88\textsubscript{$\pm$0.04} & \underline{9.95}\textsubscript{$\pm$0.06} & 47.74\textsubscript{$\pm$0.07} & 33.41\textsubscript{$\pm$0.18} & 51.09\textsubscript{$\pm$0.06} & 37.42\textsubscript{$\pm$0.30} & 29.81\textsubscript{$\pm$0.07} & 21.23\textsubscript{$\pm$0.11} & 43.62\textsubscript{$\pm$0.09} & 33.45\textsubscript{$\pm$0.15} & \underline{18.67}\textsubscript{$\pm$0.02} & \underline{13.04}\textsubscript{$\pm$0.05} & 20.36\textsubscript{$\pm$0.03} & 15.15\textsubscript{$\pm$0.05}\\
    \midrule
        SINN~\cite{okawa2022predicting} & \underline{11.95}\textsubscript{$\pm$0.04} & \underline{8.14}\textsubscript{$\pm$0.04} & \underline{13.82}\textsubscript{$\pm$0.07} & 9.98\textsubscript{$\pm$0.10} & 47.09\textsubscript{$\pm$0.09} & 35.63\textsubscript{$\pm$0.23} & \underline{50.27}\textsubscript{$\pm$0.13} & 36.58\textsubscript{$\pm$0.56} & 26.31\textsubscript{$\pm$0.15} & 17.98\textsubscript{$\pm$0.18} & 41.46\textsubscript{$\pm$0.20} & \underline{30.75}\textsubscript{$\pm$0.24} & 18.84\textsubscript{$\pm$0.04} & 13.08\textsubscript{$\pm$0.05} & 20.58\textsubscript{$\pm$0.08} & 15.17\textsubscript{$\pm$0.11}\\
        SIGNN~\cite{yang2025sociologically} & 13.43\textsubscript{$\pm$0.07} & 9.93\textsubscript{$\pm$0.12} & 14.46\textsubscript{$\pm$0.11} & 10.74\textsubscript{$\pm$0.12} & 50.95\textsubscript{$\pm$0.12} & 34.98\textsubscript{$\pm$0.30} & 53.17\textsubscript{$\pm$0.05} & 40.06\textsubscript{$\pm$0.37} & 31.75\textsubscript{$\pm$0.13} & 22.18\textsubscript{$\pm$0.20} & 46.82\textsubscript{$\pm$0.18} & 36.08\textsubscript{$\pm$0.26} & 21.97\textsubscript{$\pm$0.06} & 17.20\textsubscript{$\pm$0.07} & 24.04\textsubscript{$\pm$0.07} & 17.79\textsubscript{$\pm$0.07} \\
        ODENet~\cite{lv2023unified} & 12.56\textsubscript{$\pm$0.03} & 8.87\textsubscript{$\pm$0.05} & 14.16\textsubscript{$\pm$0.04} & 10.39\textsubscript{$\pm$0.05} & 51.63\textsubscript{$\pm$0.04} & 35.68\textsubscript{$\pm$0.14} & 53.16\textsubscript{$\pm$0.08} & 38.34\textsubscript{$\pm$0.56} & 33.25\textsubscript{$\pm$0.09} & 24.74\textsubscript{$\pm$0.10} & 45.21\textsubscript{$\pm$0.10} & 34.98\textsubscript{$\pm$0.13} & 22.04\textsubscript{$\pm$0.03} & 17.24\textsubscript{$\pm$0.03} & 24.84\textsubscript{$\pm$0.04} & 18.61\textsubscript{$\pm$0.06}\\
        GRAND~\cite{chamberlain2021grand} & 12.87\textsubscript{$\pm$0.04} & 8.99\textsubscript{$\pm$0.04} & 14.57\textsubscript{$\pm$0.03} & 10.61\textsubscript{$\pm$0.08} & 48.73\textsubscript{$\pm$0.04} & 33.57\textsubscript{$\pm$0.17} & 52.58\textsubscript{$\pm$0.05} & 37.83\textsubscript{$\pm$0.38} & 30.98\textsubscript{$\pm$0.06} & 20.16\textsubscript{$\pm$0.13} & 44.74\textsubscript{$\pm$0.08} & 34.59\textsubscript{$\pm$0.41} & 19.22\textsubscript{$\pm$0.02} & 13.51\textsubscript{$\pm$0.05} & 22.23\textsubscript{$\pm$0.05} & 16.73\textsubscript{$\pm$0.05}\\  
        GREAD~\cite{choi2023gread} & 12.21\textsubscript{$\pm$0.02} & 8.28\textsubscript{$\pm$0.03} & 13.95\textsubscript{$\pm$0.04} & 10.08\textsubscript{$\pm$0.06} & 45.95\textsubscript{$\pm$0.07} & 30.43\textsubscript{$\pm$0.20} & 51.15\textsubscript{$\pm$0.06} & \underline{35.42}\textsubscript{$\pm$0.44} & \underline{25.36}\textsubscript{$\pm$0.10} & \underline{17.86}\textsubscript{$\pm$0.17} & \underline{41.17}\textsubscript{$\pm$0.13} & 30.84\textsubscript{$\pm$0.17} & 18.78\textsubscript{$\pm$0.03} & 13.42\textsubscript{$\pm$0.07} & 21.02\textsubscript{$\pm$0.04} & 15.71\textsubscript{$\pm$0.05}\\
        AdvDifformer~\cite{wu2023advective} & 12.13\textsubscript{$\pm$0.05} & 8.21\textsubscript{$\pm$0.06} & 13.91\textsubscript{$\pm$0.04} & 10.22\textsubscript{$\pm$0.05} & 46.17\textsubscript{$\pm$0.03} & 31.52\textsubscript{$\pm$0.16} & 50.73\textsubscript{$\pm$0.09} & 36.54\textsubscript{$\pm$0.47} & 26.17\textsubscript{$\pm$0.09} & 18.42\textsubscript{$\pm$0.14} & 42.52\textsubscript{$\pm$0.16} & 31.59\textsubscript{$\pm$0.20} & 18.72\textsubscript{$\pm$0.02} & 13.28\textsubscript{$\pm$0.08} & 20.89\textsubscript{$\pm$0.06} & 15.23\textsubscript{$\pm$0.07}\\
    \midrule
        \textsc{Opinn} (Ours) & \textbf{11.70}\textsubscript{$\pm$0.02} & \textbf{7.81}\textsubscript{$\pm$0.02} & \textbf{13.65}\textsubscript{$\pm$0.03} & \textbf{9.74}\textsubscript{$\pm$0.06} & \textbf{44.43}\textsubscript{$\pm$0.05} & \textbf{27.29}\textsubscript{$\pm$0.18} & \textbf{49.52}\textsubscript{$\pm$0.04} & \textbf{34.51}\textsubscript{$\pm$0.42} & \textbf{24.76}\textsubscript{$\pm$0.11} & \textbf{16.32}\textsubscript{$\pm$0.12} & \textbf{40.49}\textsubscript{$\pm$0.16} & \textbf{30.41}\textsubscript{$\pm$0.21} & \textbf{18.26}\textsubscript{$\pm$0.02} & \textbf{12.36}\textsubscript{$\pm$0.04} & \textbf{20.05}\textsubscript{$\pm$0.05} & \textbf{14.34}\textsubscript{$\pm$0.06} \\
    \midrule
        Improvement & 2.09\% & 4.05\% & 1.23\% & 2.11\% & 2.12\% & 8.27\% & 1.55\% & 2.57\% & 2.37\% & 8.62\% & 1.65\% & 1.11\% & 2.20\% & 5.21\% & 0.79\% & 3.30\%\\
\bottomrule
\end{tabular}  
}
\end{table*}
\begin{table}[htbp]
    \centering
    \caption{Performance results on synthetic datasets ($\times 10^{-2}$). }
    \small
    \label{synresults}
    \resizebox{0.48\textwidth}{!}{
    \begin{tabular}{@{}c cc cc cc @{}} 
        \toprule
        \multirow{2}{*}{\textbf{Model}} & \multicolumn{2}{c}{\textbf{Syn-Consensus}} & \multicolumn{2}{c}{\textbf{Syn-Polarization}} & \multicolumn{2}{c}{\textbf{Syn-Clustering}}  \\
        \cmidrule(lr){2-3} \cmidrule(lr){4-5} \cmidrule(lr){6-7} 
        &RMSE & MAE & RMSE & MAE & RMSE & MAE  \\
        \midrule
        FJ~\cite{friedkin1990social} & 8.13\textsubscript{$\pm$0.00} & 6.51\textsubscript{$\pm$0.00} & 11.15\textsubscript{$\pm$0.00} & 9.20\textsubscript{$\pm$0.00} & 12.18\textsubscript{$\pm$0.00} & 10.01\textsubscript{$\pm$0.00} \\
        HK~\cite{hegselmann2015opinion} & 9.71\textsubscript{$\pm$0.00} & 8.35\textsubscript{$\pm$0.00} & 11.91\textsubscript{$\pm$0.00} & 9.33\textsubscript{$\pm$0.00} & 14.46\textsubscript{$\pm$0.00} & 11.86\textsubscript{$\pm$0.00} \\
        DeepInf~\cite{qiu2018deepinf} & 6.98\textsubscript{$\pm$0.03} & 5.26\textsubscript{$\pm$0.05} & 8.56\textsubscript{$\pm$0.02} & 7.05\textsubscript{$\pm$0.03} & 8.74\textsubscript{$\pm$0.03} & 6.19\textsubscript{$\pm$0.04} \\
        iTransformer~\cite{liu2023itransformer} & \underline{3.45}\textsubscript{$\pm$0.03} & \underline{2.41}\textsubscript{$\pm$0.04} & 5.05\textsubscript{$\pm$0.04} & 4.02\textsubscript{$\pm$0.04} & \underline{4.86}\textsubscript{$\pm$0.04} & \underline{3.52}\textsubscript{$\pm$0.04}\\
        UniGO~\cite{li2025unigo}  & 4.21\textsubscript{$\pm$0.02} & 2.99\textsubscript{$\pm$0.03} & \underline{4.98}\textsubscript{$\pm$0.02} & \underline{3.89}\textsubscript{$\pm$0.03} & 5.07\textsubscript{$\pm$0.01} & 3.65\textsubscript{$\pm$0.02} \\
        SINN~\cite{okawa2022predicting} & 4.42\textsubscript{$\pm$0.04} & 3.35\textsubscript{$\pm$0.05} & 5.47\textsubscript{$\pm$0.02} & 4.23\textsubscript{$\pm$0.05} & 5.88\textsubscript{$\pm$0.03} & 4.43\textsubscript{$\pm$0.03} \\
        GRAND~\cite{chamberlain2021grand} & 4.14\textsubscript{$\pm$0.01} & 3.11\textsubscript{$\pm$0.01} & 5.69\textsubscript{$\pm$0.01} & 4.40\textsubscript{$\pm$0.02} & 5.98\textsubscript{$\pm$0.01} & 4.44\textsubscript{$\pm$0.03}\\
        GREAD~\cite{choi2023gread} & 4.07\textsubscript{$\pm$0.01} & 3.06\textsubscript{$\pm$0.02} & 5.54\textsubscript{$\pm$0.02} & 4.24\textsubscript{$\pm$0.02} & 5.66\textsubscript{$\pm$0.02} & 4.21\textsubscript{$\pm$0.04}\\
        AdvDifformer~\cite{wu2023advective} & 3.97\textsubscript{$\pm$0.02} & 2.97\textsubscript{$\pm$0.03} & 5.12\textsubscript{$\pm$0.03} & 4.08\textsubscript{$\pm$0.04} & 5.11\textsubscript{$\pm$0.03} & 4.09\textsubscript{$\pm$0.04}\\
        \midrule
        \textsc{Opinn} (Ours) & \textbf{3.29}\textsubscript{$\pm$0.01} & \textbf{2.35}\textsubscript{$\pm$0.02} & \textbf{4.46}\textsubscript{$\pm$0.03} & \textbf{3.44}\textsubscript{$\pm$0.04} & \textbf{4.60}\textsubscript{$\pm$0.02} & \textbf{3.35}\textsubscript{$\pm$0.03} \\
        \midrule
        Improvement & 4.64\% & 2.49\% & 10.44\% & 11.57\% & 5.35\% & 4.83\%\\
        \bottomrule
    \end{tabular}
    }
\end{table}
\begin{table}[htbp]
    \centering
    \caption{Performance results under few-shot setting ($\times 10^{-2}$).}
    \small
    \label{limitdata}
    \resizebox{0.49\textwidth}{!}{
    \begin{tabular}{@{}c cc cc cc cc@{}} 
        \toprule
        \multirow{2}{*}{\textbf{Model}} & \multicolumn{2}{c}{\textbf{Delhi Election}} & \multicolumn{2}{c}{\textbf{U.S. Election}} & \multicolumn{2}{c}{\textbf{Israel-Palestine}} & \multicolumn{2}{c}{\textbf{COVID-19}} \\
        \cmidrule(lr){2-3} \cmidrule(lr){4-5} \cmidrule(lr){6-7} \cmidrule(lr){8-9}
        &RMSE & MAE & RMSE & MAE & RMSE & MAE & RMSE & MAE \\
        \midrule
        SGFormer~\cite{wu2023sgformer} & 13.79\textsubscript{$\pm$0.07} & 9.84\textsubscript{$\pm$0.08} & 51.13\textsubscript{$\pm$0.12} & 36.70\textsubscript{$\pm$0.26} & 40.05\textsubscript{$\pm$0.25} & 28.12\textsubscript{$\pm$0.30} & 20.02\textsubscript{$\pm$0.07} & 13.93\textsubscript{$\pm$0.10}\\
        iTransformer~\cite{liu2023itransformer} & 12.46\textsubscript{$\pm$0.04} & 8.71\textsubscript{$\pm$0.06} & 48.59\textsubscript{$\pm$0.10} & 35.25\textsubscript{$\pm$0.23} & 38.63\textsubscript{$\pm$0.27}  & 26.55\textsubscript{$\pm$0.29}  & 19.03\textsubscript{$\pm$0.06} & 13.35\textsubscript{$\pm$0.07}\\
        UniGO~\cite{li2025unigo} & \underline{12.19}\textsubscript{$\pm$0.05} & \underline{8.32}\textsubscript{$\pm$0.08} & 46.28\textsubscript{$\pm$0.15} & 29.94\textsubscript{$\pm$0.37} & 37.85\textsubscript{$\pm$0.29} & 25.86\textsubscript{$\pm$0.36} & 19.77\textsubscript{$\pm$0.10} & 14.25\textsubscript{$\pm$0.13} \\
        SINN~\cite{okawa2022predicting}  & 12.37\textsubscript{$\pm$0.07} & 8.57\textsubscript{$\pm$0.09} & \underline{46.14}\textsubscript{$\pm$0.14} & \underline{29.80}\textsubscript{$\pm$0.24} & 38.12\textsubscript{$\pm$0.21} & 25.98\textsubscript{$\pm$0.30} & \underline{18.97}\textsubscript{$\pm$0.09} & \underline{13.02}\textsubscript{$\pm$0.09} \\
        GREAD~\cite{choi2023gread} & 12.42\textsubscript{$\pm$0.06} & 8.69\textsubscript{$\pm$0.06} & 47.84\textsubscript{$\pm$0.07} & 33.43\textsubscript{$\pm$0.22} & 37.23\textsubscript{$\pm$0.28} & 25.67\textsubscript{$\pm$0.37} & 19.42\textsubscript{$\pm$0.05} & 13.69\textsubscript{$\pm$0.06}\\
        AdvDifformer~\cite{wu2023advective} & 12.51\textsubscript{$\pm$0.06} & 8.87\textsubscript{$\pm$0.10} & 47.53\textsubscript{$\pm$0.05} & 32.26\textsubscript{$\pm$0.20} & \underline{37.07}\textsubscript{$\pm$0.25} & \underline{25.23}\textsubscript{$\pm$0.34} & 19.31\textsubscript{$\pm$0.07} & 13.47\textsubscript{$\pm$0.08}\\
        \midrule
        \textsc{Opinn} (Ours) & \textbf{11.97}\textsubscript{$\pm$0.05} & \textbf{7.96}\textsubscript{$\pm$0.07} & \textbf{44.73}\textsubscript{$\pm$0.08} & \textbf{28.97}\textsubscript{$\pm$0.21} & \textbf{35.46}\textsubscript{$\pm$0.23} & \textbf{23.06}\textsubscript{$\pm$0.31} & \textbf{18.68}\textsubscript{$\pm$0.08} & \textbf{12.59}\textsubscript{$\pm$0.09} \\
        \bottomrule
    \end{tabular}
    }
\end{table}
\subsection{Performance Comparison}
\noindent \textbf{Real-world datasets.}
We compare \textsc{Opinn} with 16 representative baselines on real-world datasets. 
Experimental results with standard deviations are shown in Table~\ref{rearesults}, and we can observe:
(1) \textsc{Opinn} consistently outperforms baselines on real-world datasets, achieving a notable 8.62\% reduction in MAE over the best-performing baseline on Israel-Palestine.
(2) Data-driven approaches, particularly iTransformer and UniGO, generally outperform mechanical methods. However, GCN and GAT perform even worse than the FJ model on Delhi Election, indicating that purely data-centric paradigm may result in inadequate inductive bias.
(3) Competitive performance of physic-informed methods such as SINN and GREAD validates the integration of physical priors into neural networks.
(4) The sub-optimal performance of SIGNN reveals a possible mismatch between physical and neural components, whereas the superiority of GREAD and AdvDifformer over GRAND validates the necessity of comprehensive physical priors (see analysis in Appendix~\ref{appbaseline2}).

\noindent \textbf{Synthetic datasets.}
We compare \textsc{Opinn} with 9 highly competitive baselines on synthetic datasets, and experimental results are presented in Table~\ref{synresults}.
Notably, \textsc{Opinn} retains its lead in performance, and the global aggregation of iTransformer effectively identifies the distinct evolutionary trends within synthetic datasets.

\noindent \textbf{Few-shot setting.}
We further assess the model generalization under few-shot setting to capitalize on the data efficiency of physics-informed methods.
We split real-world datasets chronologically into training, validation, and testing sets using a 30\%: 10\%: 60\% ratio, with the results shown in Table~\ref{limitdata}.
\textsc{Opinn} maintains a clear performance edge over other baselines, validating its superior capacity for opinion forecasting under significant data constraints.

\begin{figure}[htbp]
    \centering 
    \includegraphics[width=0.45\textwidth]{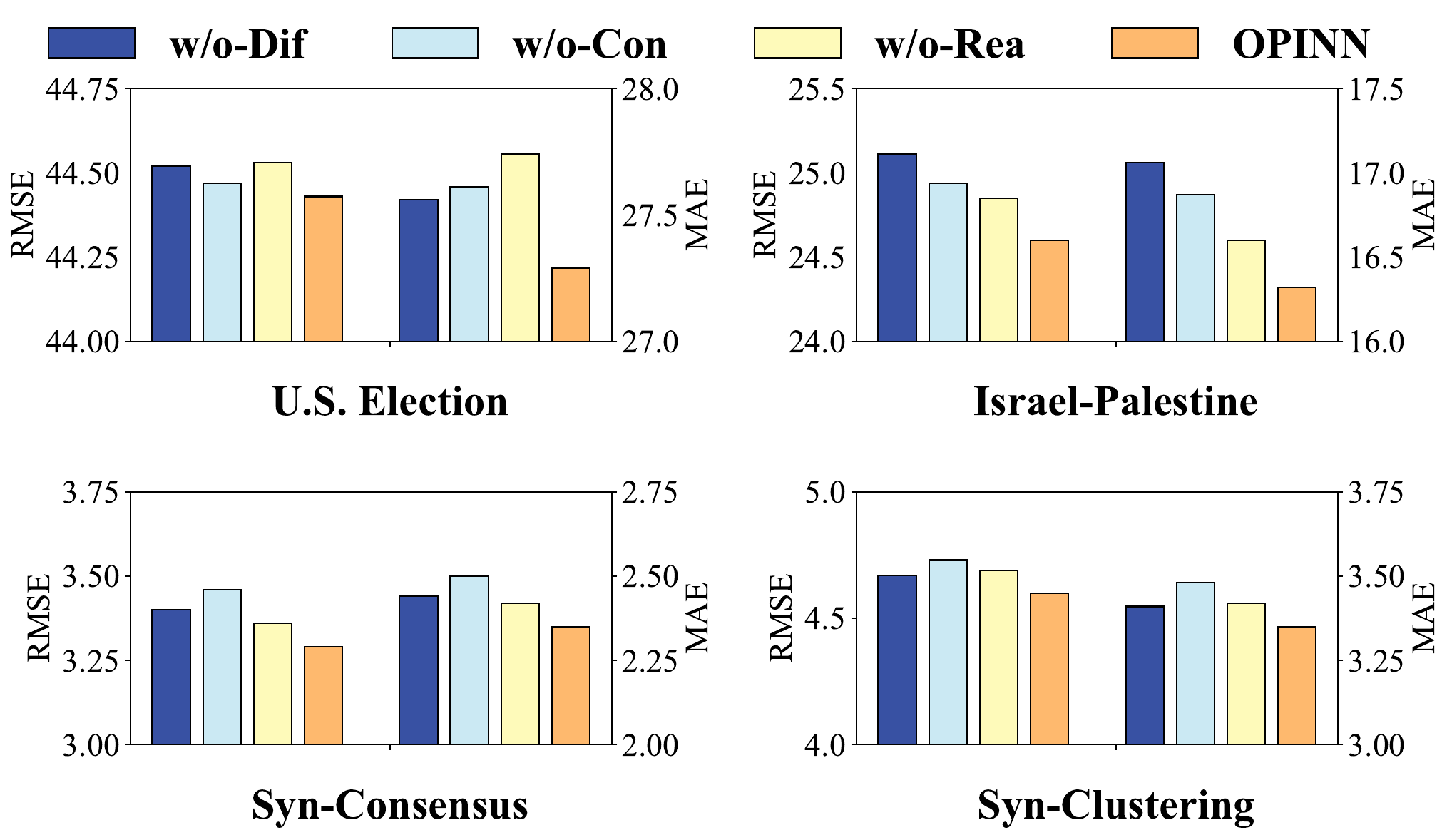} 
    \caption{Ablation results of opinion dynamics mechanisms.} 
    \label{abpro}
\end{figure}
\begin{figure}[htbp]
    \centering 
    \includegraphics[width=0.45\textwidth]{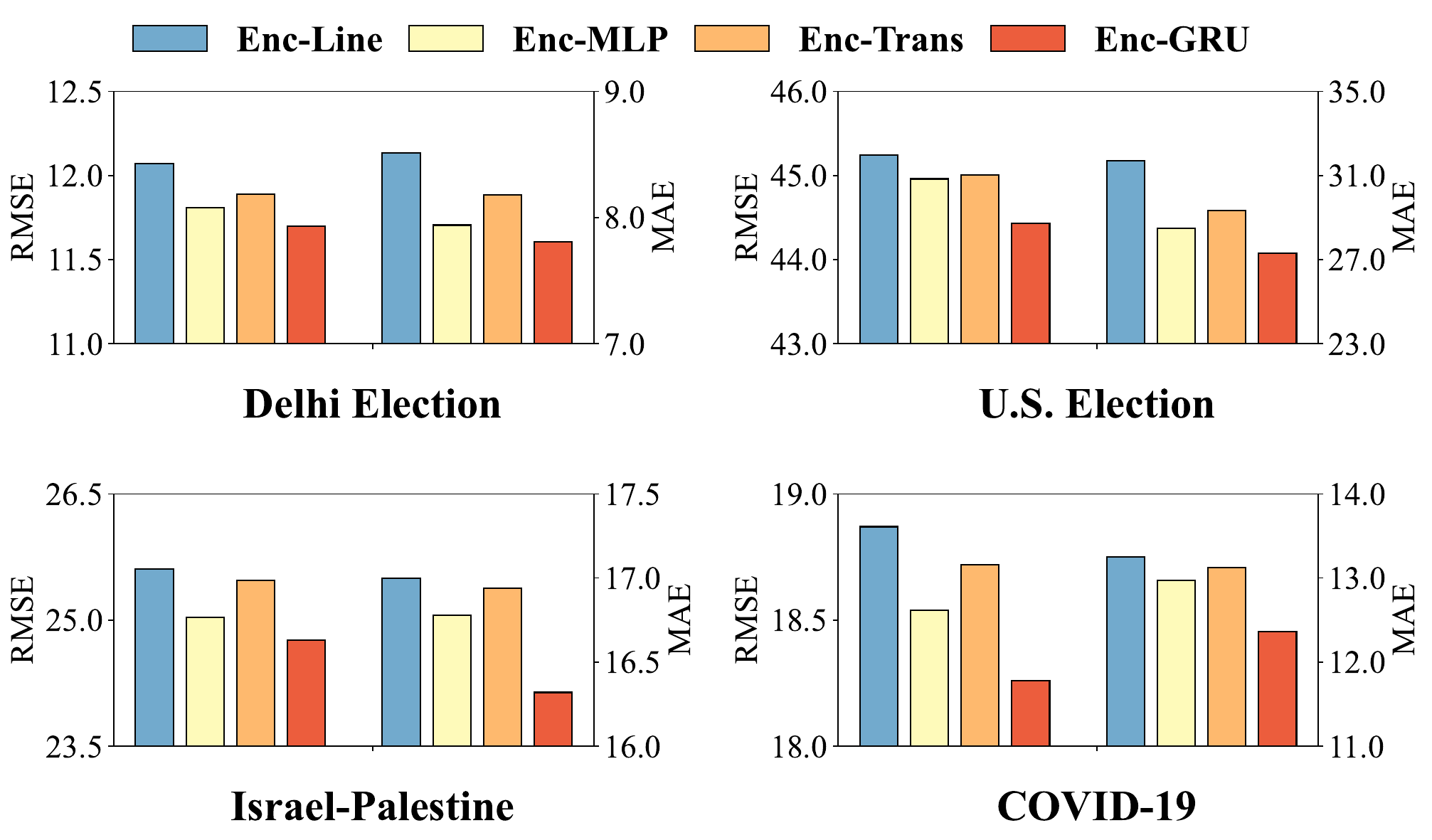} 
    \caption{The ablation study of different encoders. } 
    \label{abenc}
\end{figure}
\begin{table}[htbp]
    \centering
    \caption{Model analysis on the reaction term ($\times 10^{-2}$).}
    \small
    \label{reaction}
    \resizebox{0.45\textwidth}{!}{
    \begin{tabular}{@{}c cc cc cc cc@{}} 
        \toprule
        \multirow{2}{*}{\textbf{Reaction term}} & \multicolumn{2}{c}{\textbf{U.S. Election}} & \multicolumn{2}{c}{\textbf{COVID-19}} & \multicolumn{2}{c}{\textbf{Syn-Consensus}} & \multicolumn{2}{c}{\textbf{Syn-Polarization}} \\
        \cmidrule(lr){2-3} \cmidrule(lr){4-5} \cmidrule(lr){6-7} \cmidrule(lr){8-9}
        &RMSE & MAE & RMSE & MAE & RMSE & MAE & RMSE & MAE \\
        \midrule
        Source term & 45.37 & 28.98 & \underline{18.89} & \underline{13.00} & 3.45 & 2.47 & 4.73 & 3.69 \\
        Linear term & \underline{45.21} & \underline{28.84} & 19.25 & 13.28 & \textbf{3.29} & \textbf{2.35} & \underline{4.58} & \underline{3.59} \\
        Non-linear term & \textbf{44.43} & \textbf{27.29} & \textbf{18.26} & \textbf{12.36} & \underline{3.36} & \underline{2.39} & \textbf{4.46} & \textbf{3.44} \\
        \bottomrule
    \end{tabular}
    }
\end{table}

\subsection{Ablation Study}
\label{ablationex}
\noindent \textbf{Opinion dynamics mechanisms.}
To quantify the contribution of three opinion dynamics mechanisms, we compare \textsc{Opinn} against three variants:
\textbf{w/o-Dif}: adopt convection and reaction as priors. 
\textbf{w/o-Con}: adopt diffusion and reaction as priors.
\textbf{w/o-Rea}: adopt diffusion and convection as priors.
Figure~\ref{abpro} shows the ablation results on real-world and synthetic datasets, we observe the following:
(1) \textsc{Opinn} outperforms other variants, highlighting the efficacy of comprehensive dynamics priors.
(2) Due to the interplay of local, global, and endogenous impact, the absence of any constituent dynamics process compromises overall performance on real-world datasets.
(3) For synthetic datasets, omitting convection results in the significant performance drop due to the dominant global trends within datasets.
For additional ablation results and analysis about gating weights, please refer to Appendix~\ref{appodm}

\noindent \textbf{Encoder impact.}
To study the efficacy of different encoders, we implement \textsc{Opinn} with four variants:
\textbf{Enc-Line}: adopt a single linear layer as encoder.
\textbf{Enc-MLP}: adopt a two-layer MLP as encoder.
\textbf{Enc-Trans}: adopt a Transformer~\cite{vaswani2017attention} block as encoder.
\textbf{Enc-GRU}: adopt a Gated Recurrent Unit~\cite{chung2014empirical} as encoder.
Figure~\ref{abenc} illustrates the ablation results on real-world datasets, where GRU consistently outperforms other encoders.
Therefore, we employ GRU within \textsc{Opinn} to encode the temporal patterns of user opinions as the system state.
Notably, the Transformer-based encoder underperforms, likely because the restricted look-back window limits its inherent capacity for long-range dependency modeling.

\begin{table}[htbp]
    \centering
    \caption{Performance comparison of different ODE solvers.}
    \small
    \label{solvers}
    \resizebox{0.47\textwidth}{!}{
    \begin{tabular}{@{}c ccc ccc ccc @{}} 
        \toprule
        \multirow{2}{*}{\textbf{ODE Solver}} & \multicolumn{3}{c}{\textbf{Delhi Election}} & \multicolumn{3}{c}{\textbf{U.S. Election}} & \multicolumn{3}{c}{\textbf{COVID-19}} \\
        \cmidrule(lr){2-4} \cmidrule(lr){5-7} \cmidrule(lr){8-10} 
        &RMSE & MAE & Time (ms) & RMSE & MAE & Time (ms) & RMSE & MAE & Time (ms) \\
        \midrule
        Euler   & 11.76 & 7.85 & 19.46 & 44.54 & 28.62 & 13.52 & 18.34 & 12.37 & 22.44\\
        RK-4     & 11.70 & 7.81 & 49.76 & 44.43 & 27.29 & 34.38 & 18.26 & 12.36 & 61.52 \\
        Dopri-5  & 11.68 & 7.80 & 198.22 & 44.43 & 27.25 & 179.25 & 18.23 & 12.31 & 235.45 \\
        \bottomrule
    \end{tabular}
    }
\end{table}
\begin{figure}[htbp]
    \centering
    \includegraphics[width=0.38\textwidth]{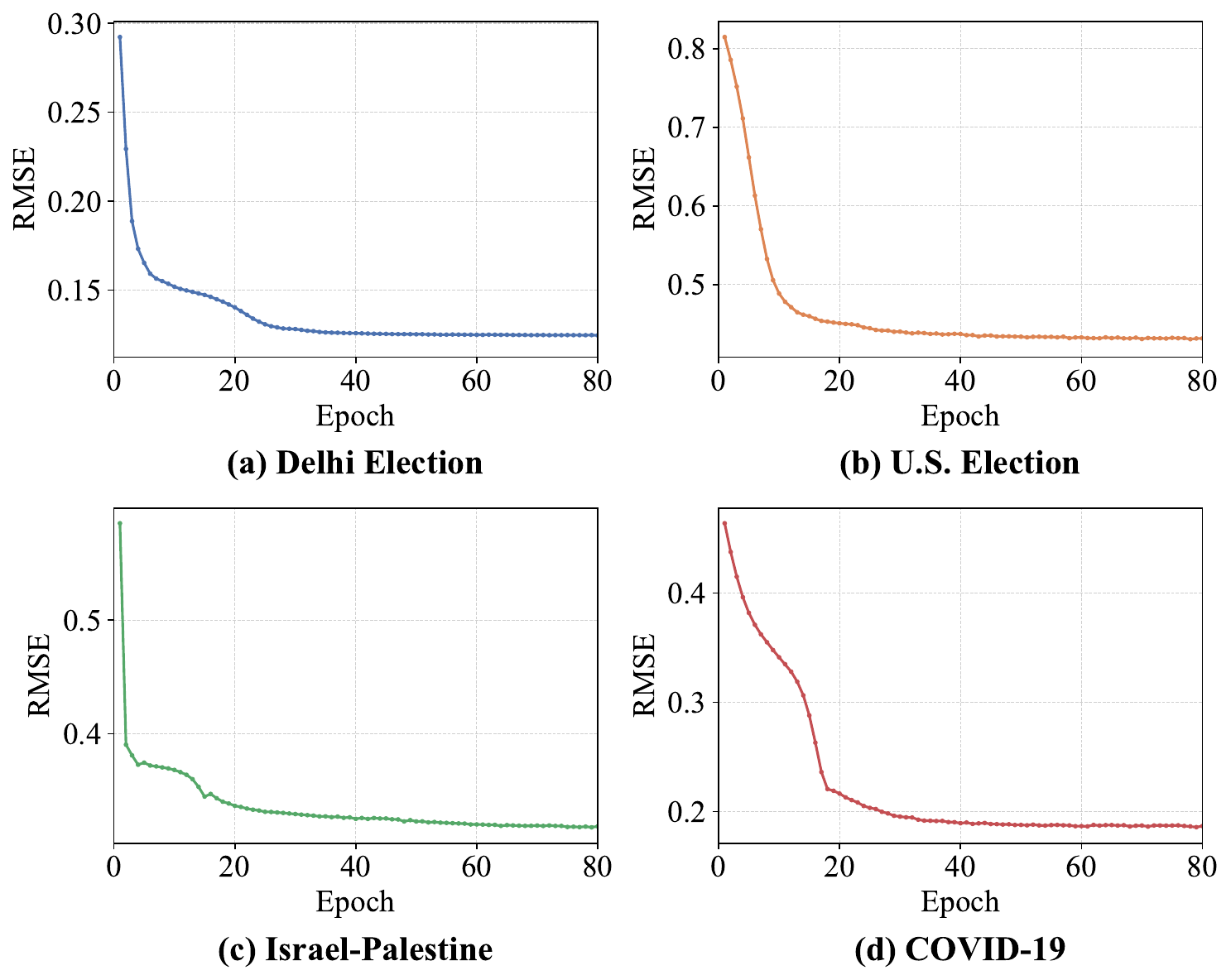} 
    \caption{The training loss of \textsc{Opinn} on real-world datasets.} 
    \label{stb} 
\end{figure}

\subsection{Model Analysis}
\label{ma}
\noindent \textbf{Reaction term.}
To investigate the impact of reaction terms, we conducted analysis studies on both real-world and synthetic datasets.
As illustrated in Table~\ref{reaction}, the nonlinear reaction exhibits superior performance on real-world datasets and Syn-Polarization, effectively capturing the intricate endogenous patterns in practical scenarios. 
Conversely, the linear term excels on Syn-Consensus datasets, indicating the consensus formation follow predominantly linear and predictable underlying dynamics. 
For additional results and analysis of the reaction term, please refer to Appendix~\ref{apprt}.

\noindent \textbf{Model complexity.}
The major time complexity of \textsc{Opinn} comes from the neural dynamics module. 
The time complexity is $O(ND^2+|\mathcal{E}|D)$ for diffusion and $O(ND^2+DN^2)$ for convection, where $|\mathcal{E}|$ is the number of edges.
Given the quadratic complexity of convection, we provide supplementary experiments in Appendix~\ref{appmc} for \textsc{Opinn} variants utilizing vanilla and linear attention scheme. 
\textsc{Opinn} achieves a competitive trade-off where its superior performance more than compensates for the slight computation overhead.
Meanwhile, we report the model sizes and computation times of \textsc{Opinn} in comparison with representative baselines.

\noindent \textbf{Numerical solver.}
To evaluate the computational stability and runtime differences of ODE solver, we implement \textsc{Opinn} via three numerical solvers: Euler, RK-4, and Dopri-5.
Table~\ref{solvers} shows the performance comparison of different ODE solvers, with running time measured at the inference phase.
We observe that high-order solvers such as RK-4 and Dopri-5 outperform the lower-order method Euler, but at the cost of higher computational overhead.
This demonstrates the trade-off between performance and solver complexity. 
In our implementation, we adopt the RK-4 as ODE solver since it can maintain acceptable computation time while ensuring accuracy.

\noindent \textbf{Numerical stability.}
Given the complex integration of an autoencoder architecture and a neural dynamics module, we evaluate the numerical stability of \textsc{Opinn}.
Figure~\ref{stb} illustrates the training loss curves on four real-world datasets, and
it is evident that~\textsc{Opinn}’s loss rapidly declines and converges during the training phase.

\begin{figure}[htbp]
    \centering
    \begin{subfigure}{0.23\textwidth}
        \centering
        \includegraphics[width=\textwidth]{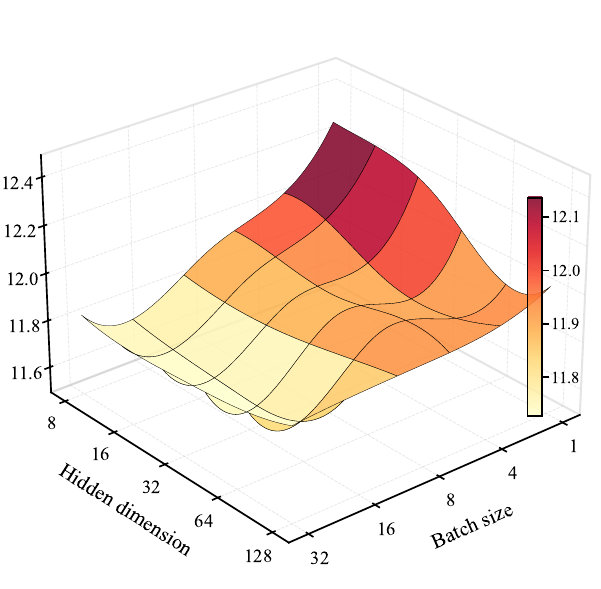} 
        \caption{Delhi Election}
    \end{subfigure}
    \begin{subfigure}{0.23\textwidth}
        \centering
        \includegraphics[width=\textwidth]{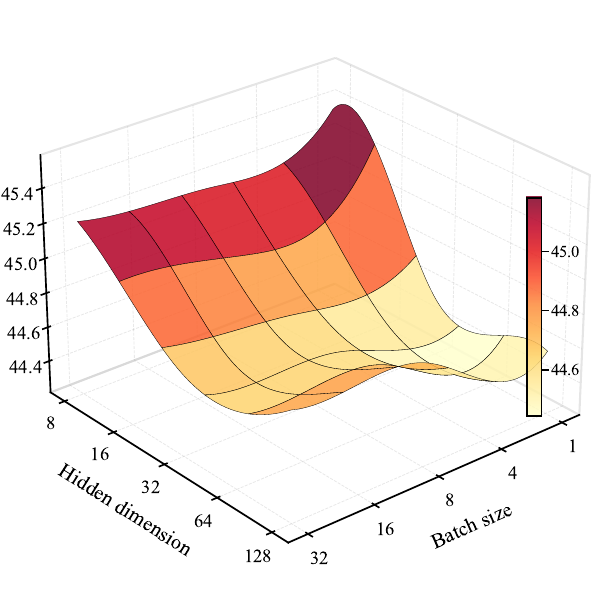} 
        \caption{U.S. Election}
    \end{subfigure}
    \caption{The hyperparameter sensitivity of hidden dimension and batch size for \textsc{Opinn} (measured by RMSE).} 
    \label{hs}
\end{figure}

\noindent \textbf{Hyperparameter sensitivity.}
We assess the hyperparameter sensitivity by analyzing model performance under different configurations of encoder hidden dimensions and batch sizes, as shown in Figure~\ref{hs}.
Results show that performance drops when hidden dimensions and batch sizes are too small. 
For moderate-scale dimensions (32, 64), we find that a smaller batch size (4) benefits U.S. Election, while Delhi Election requires a larger batch size (16).

\begin{figure}[htbp]
    \centering
    \begin{subfigure}{0.49\textwidth}
        \centering
        \includegraphics[width=\textwidth]{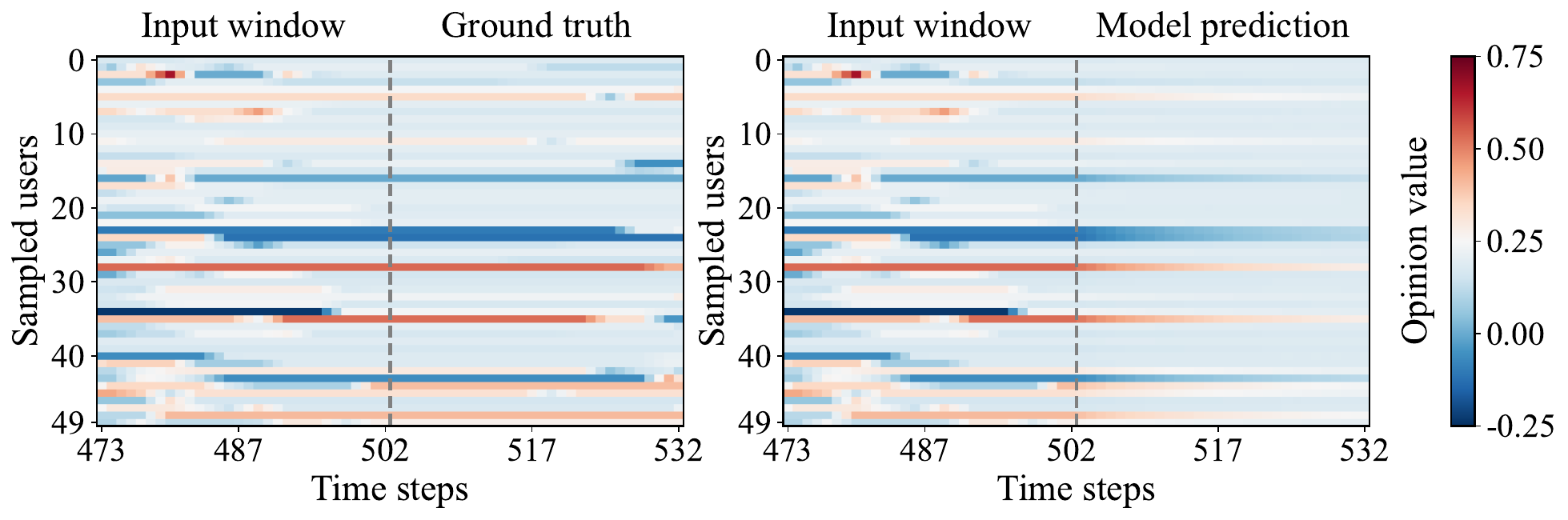} 
        \caption{Visualization of opinions for 50 randomly sampled users.}
        \label{cs1}
    \end{subfigure}
    \begin{subfigure}{0.49\textwidth}
        \centering
        \includegraphics[width=\textwidth]{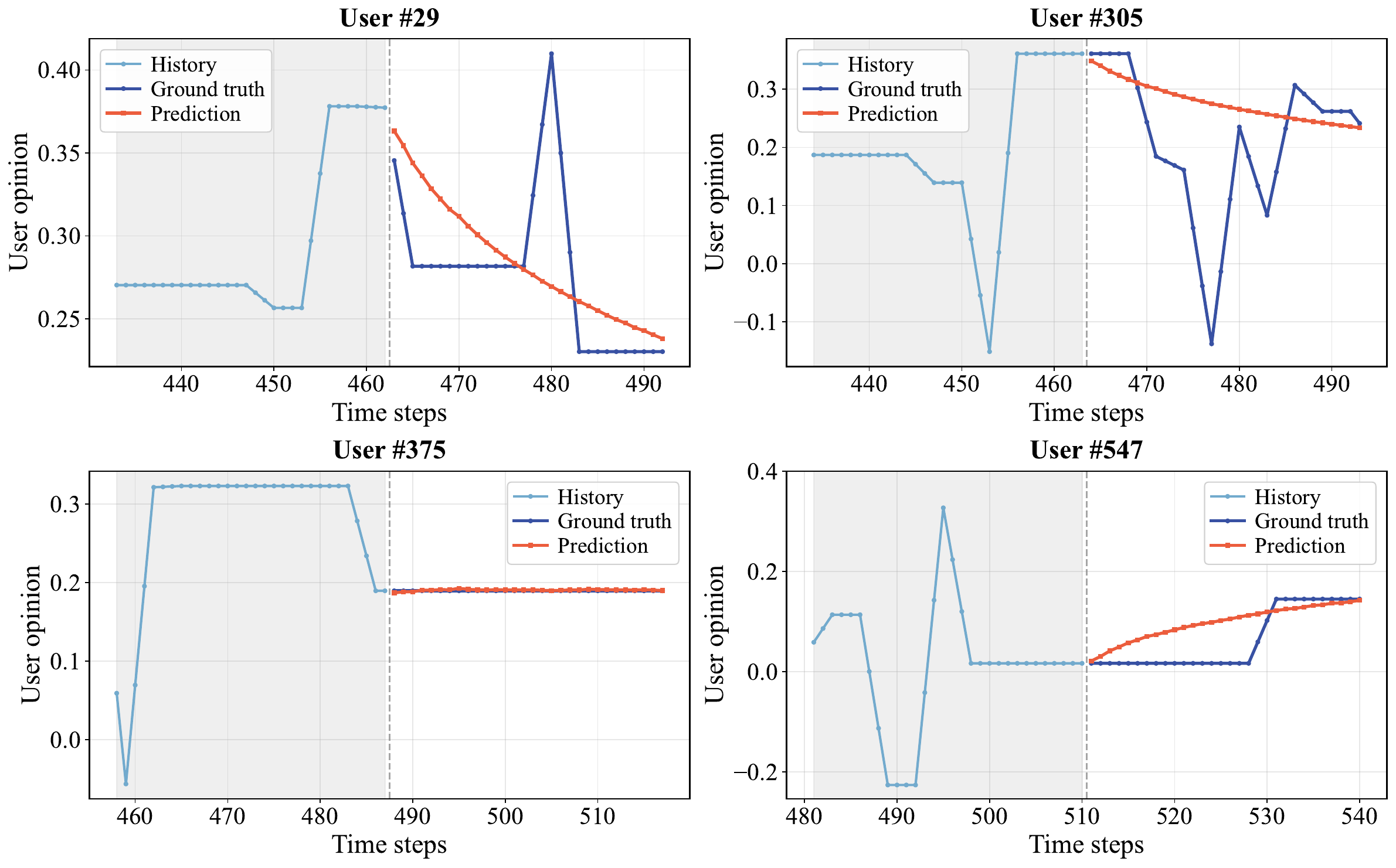} 
        \caption{Visualization of opinions for 4 typical users. }
        \label{cs2}
    \end{subfigure}
    \caption{Case studies on the Delhi Election dataset.} 
\end{figure}

\subsection{Case Study}
To evaluate the predictive performance, we conduct case studies at the macro and micro levels on real-world datasets.
As shown in Figure~\ref{cs1}, we present the ground truth versus model predictions of 50 randomly sampled users from the Delhi Election dataset, indicating that our model can accurately track the overall macro-level dynamics.
At the micro level, we visualize the opinion evolution of 4 typical users in Figure~\ref{cs2}.
The findings reveal that \textsc{Opinn} remains robust and provides precise predictions, despite the inherent noise of real-world social platforms.
For more case studies on other datasets, please refer to Appendix~\ref{appcs}.

\section{Conclusion}
This paper presents \textsc{Opinn}, a physics-informed framework for opinion dynamics modeling.
By introducing the DCR system, we provide a comprehensive physical system to interpret real-world opinion dynamics. 
Inspired by Neural ODEs, we define the neural opinion dynamics to effectively synergize neural representations with physical priors.
Extensive experiments have proven that our proposed \textsc{Opinn} achieves superior performance and generalization for real-world opinion dynamics forecasting.

\section{Limitation \& Future Work}
\noindent \textbf{Limited features}.
In this study, user features are limited to observed opinions derived from LLMs and manual assessment. Future work can expand this to include broader attributes, including demographic information (e.g., age and gender) and personal interests.

\noindent \textbf{Computation bottleneck.} 
The convection process in \textsc{Opinn} entails a quadratic complexity relative to the number of uses.
While techniques like user sampling and linearized attention (see our discussions in Appendix~\ref{appcc}) can alleviate this issue, the computational bottleneck limits the model's scalability to large-scale networks.

\noindent \textbf{Opinion simulation.} 
Given the challenges in collecting opinion data and social networks, leveraging the powerful LLMs for agent-based simulation for trending issues presents a promising direction.
Such paradigm can not only infer the evolution of broader topics but also provide comprehensive benchmarks for model evaluation.

\newpage


\bibliographystyle{ACM-Reference-Format}
\bibliography{ref}

@String{Computing = "Computing" }

@String{Computer = "{IEEE} Computer" }

@String{Springer = "Springer-Verlag" }

@article{kipf2016semi,
  title={Semi-supervised classification with graph convolutional networks},
  author={Kipf, TN},
  journal={arXiv preprint arXiv:1609.02907},
  year={2016}
}

@article{velivckovic2017graph,
  title={Graph attention networks},
  author={Veli{\v{c}}kovi{\'c}, Petar and Cucurull, Guillem and Casanova, Arantxa and Romero, Adriana and Lio, Pietro and Bengio, Yoshua},
  journal={arXiv preprint arXiv:1710.10903},
  year={2017}
}

@inproceedings{yildiz2010voting,
  title={Voting models in random networks},
  author={Yildiz, Mehmet E and Pagliari, Roberto and Ozdaglar, Asuman and Scaglione, Anna},
  booktitle={2010 information theory and applications workshop (ITA)},
  pages={1--7},
  year={2010},
  organization={IEEE}
}

@article{degroot1974reaching,
  title={Reaching a consensus},
  author={DeGroot, Morris H},
  journal={Journal of the American Statistical association},
  volume={69},
  number={345},
  pages={118--121},
  year={1974},
  publisher={Taylor \& Francis}
}

@inproceedings{okawa2022predicting,
  title={Predicting opinion dynamics via sociologically-informed neural networks},
  author={Okawa, Maya and Iwata, Tomoharu},
  booktitle={Proceedings of the 28th ACM SIGKDD conference on knowledge discovery and data mining},
  pages={1306--1316},
  year={2022}
}

@inproceedings{yang2025sociologically,
  title={Sociologically-Informed Graph Neural Network for Opinion Prediction},
  author={Yang, Fan and Bai, Jie and Li, Linjing and Zeng, Daniel},
  booktitle={ICASSP 2025-2025 IEEE International Conference on Acoustics, Speech and Signal Processing (ICASSP)},
  pages={1--5},
  year={2025},
  organization={IEEE}
}

@inproceedings{li2025unigo,
  title={UniGO: A Unified Graph Neural Network for Modeling Opinion Dynamics on Graphs},
  author={Li, Hao and Jiang, Hao and Zheng, Yuke and Sun, Hao and Gong, Wenying},
  booktitle={Proceedings of the ACM on Web Conference 2025},
  pages={530--540},
  year={2025}
}

@inproceedings{ye2012exploring,
  title={Exploring social influence for recommendation: a generative model approach},
  author={Ye, Mao and Liu, Xingjie and Lee, Wang-Chien},
  booktitle={Proceedings of the 35th international ACM SIGIR conference on Research and development in information retrieval},
  pages={671--680},
  year={2012}
}

@inproceedings{lai2018stance,
  title={Stance evolution and twitter interactions in an italian political debate},
  author={Lai, Mirko and Patti, Viviana and Ruffo, Giancarlo and Rosso, Paolo},
  booktitle={International conference on applications of natural language to information systems},
  pages={15--27},
  year={2018},
  organization={Springer}
}

@article{bond201261,
  title={A 61-million-person experiment in social influence and political mobilization},
  author={Bond, Robert M and Fariss, Christopher J and Jones, Jason J and Kramer, Adam DI and Marlow, Cameron and Settle, Jaime E and Fowler, James H},
  journal={Nature},
  volume={489},
  number={7415},
  pages={295--298},
  year={2012},
  publisher={Nature Publishing Group UK London}
}

@article{santos2021link,
  title={Link recommendation algorithms and dynamics of polarization in online social networks},
  author={Santos, Fernando P and Lelkes, Yphtach and Levin, Simon A},
  journal={Proceedings of the National Academy of Sciences},
  volume={118},
  number={50},
  pages={e2102141118},
  year={2021},
  publisher={National Academy of Sciences}
}

@inproceedings{o2010tweets,
  title={From tweets to polls: Linking text sentiment to public opinion time series},
  author={O'Connor, Brendan and Balasubramanyan, Ramnath and Routledge, Bryan and Smith, Noah},
  booktitle={Proceedings of the international AAAI conference on web and social media},
  volume={4},
  number={1},
  pages={122--129},
  year={2010}
}

@article{lv2023unified,
  title={A unified view on neural message passing with opinion dynamics for social networks},
  author={Lv, Outongyi and Zhou, Bingxin and Wang, Jing and Xiao, Xiang and Zhao, Weishu and Zheng, Lirong},
  journal={arXiv preprint arXiv:2310.01272},
  year={2023}
}

@article{chung2014empirical,
  title={Empirical evaluation of gated recurrent neural networks on sequence modeling},
  author={Chung, Junyoung and Gulcehre, Caglar and Cho, KyungHyun and Bengio, Yoshua},
  journal={arXiv preprint arXiv:1412.3555},
  year={2014}
}

@article{friedkin1990social,
  title={Social influence and opinions},
  author={Friedkin, Noah E and Johnsen, Eugene C},
  journal={Journal of mathematical sociology},
  volume={15},
  number={3-4},
  pages={193--206},
  year={1990},
  publisher={Taylor \& Francis}
}

@article{hegselmann2015opinion,
  title={Opinion dynamics and bounded confidence: models, analysis and simulation},
  author={Hegselmann, Rainer},
  journal={The Journal of Artificial Societies and Social Simulation},
  year={2015}
}

@article{sznajd2000opinion,
  title={Opinion evolution in closed community},
  author={Sznajd-Weron, Katarzyna and Sznajd, Jozef},
  journal={International Journal of Modern Physics C},
  volume={11},
  number={06},
  pages={1157--1165},
  year={2000},
  publisher={World Scientific}
}

@article{raissi2019physics,
  title={Physics-informed neural networks: A deep learning framework for solving forward and inverse problems involving nonlinear partial differential equations},
  author={Raissi, Maziar and Perdikaris, Paris and Karniadakis, George E},
  journal={Journal of Computational physics},
  volume={378},
  pages={686--707},
  year={2019},
  publisher={Elsevier}
}

@article{liu2023itransformer,
  title={itransformer: Inverted transformers are effective for time series forecasting},
  author={Liu, Yong and Hu, Tengge and Zhang, Haoran and Wu, Haixu and Wang, Shiyu and Ma, Lintao and Long, Mingsheng},
  journal={arXiv preprint arXiv:2310.06625},
  year={2023}
}

@inproceedings{chamberlain2021grand,
  title={Grand: Graph neural diffusion},
  author={Chamberlain, Ben and Rowbottom, James and Gorinova, Maria I and Bronstein, Michael and Webb, Stefan and Rossi, Emanuele},
  booktitle={International conference on machine learning},
  pages={1407--1418},
  year={2021},
  organization={PMLR}
}

@inproceedings{qiu2018deepinf,
  title={Deepinf: Social influence prediction with deep learning},
  author={Qiu, Jiezhong and Tang, Jian and Ma, Hao and Dong, Yuxiao and Wang, Kuansan and Tang, Jie},
  booktitle={Proceedings of the 24th ACM SIGKDD international conference on knowledge discovery \& data mining},
  pages={2110--2119},
  year={2018}
}

@article{wu2023sgformer,
  title={Sgformer: Simplifying and empowering transformers for large-graph representations},
  author={Wu, Qitian and Zhao, Wentao and Yang, Chenxiao and Zhang, Hengrui and Nie, Fan and Jiang, Haitian and Bian, Yatao and Yan, Junchi},
  journal={Advances in Neural Information Processing Systems},
  volume={36},
  pages={64753--64773},
  year={2023}
}

@article{shirzadi2025opinion,
  title={Opinion Dynamics: A Comprehensive Overview},
  author={Shirzadi, Mohammad and Cruciani, Emilio and Zehmakan, Ahad N},
  journal={arXiv preprint arXiv:2511.00401},
  year={2025}
}

@article{anstead2015social,
  title={Social media analysis and public opinion: The 2010 UK general election},
  author={Anstead, Nick and O'Loughlin, Ben},
  journal={Journal of computer-mediated communication},
  volume={20},
  number={2},
  pages={204--220},
  year={2015},
  publisher={Oxford University Press Oxford, UK}
}

@inproceedings{duan2025bi,
  title={Bi-Dynamic Graph ODE for Opinion Evolution},
  author={Duan, Bowen and Deng, Henggang and Piao, Jinghua and Wang, Huandong and Wang, Yue},
  booktitle={Proceedings of the 31st ACM SIGKDD Conference on Knowledge Discovery and Data Mining V. 1},
  pages={260--270},
  year={2025}
}

@article{barrio2015dynamics,
  title={Dynamics of deceptive interactions in social networks},
  author={Barrio, Rafael A and Govezensky, Tzipe and Dunbar, Robin and Iniguez, Gerardo and Kaski, Kimmo},
  journal={Journal of the Royal society Interface},
  volume={12},
  number={112},
  pages={20150798},
  year={2015},
  publisher={The Royal Society}
}

@article{chandrasekaran2025network,
  title={Network-aware recommender system via online feedback optimization},
  author={Chandrasekaran, Sanjay and De Pasquale, Giulia and Belgioioso, Giuseppe and D{\"o}rfler, Florian},
  journal={IEEE Transactions on Automatic Control},
  year={2025},
  publisher={IEEE}
}

@article{de2019learning,
  title={Learning linear influence models in social networks from transient opinion dynamics},
  author={De, Abir and Bhattacharya, Sourangshu and Bhattacharya, Parantapa and Ganguly, Niloy and Chakrabarti, Soumen},
  journal={ACM Transactions on the Web (TWEB)},
  volume={13},
  number={3},
  pages={1--33},
  year={2019},
  publisher={ACM New York, NY, USA}
}

@article{hou2021opinion,
  title={Opinion dynamics in modified expressed and private model with bounded confidence},
  author={Hou, Jian and Li, Wenshan and Jiang, Mingyue},
  journal={Physica A: Statistical Mechanics and its Applications},
  volume={574},
  pages={125968},
  year={2021},
  publisher={Elsevier}
}

@article{gong2026survey,
  title={A survey on learning from graphs with heterophily: Recent advances and future directions},
  author={Gong, Cheng-Hua and Cheng, Yao and Yu, Jian-Xiang and Xu, Can and Shan, Cai-Hua and Luo, Si-Qiang and Li, Xiang},
  journal={Frontiers of Computer Science},
  volume={20},
  number={2},
  pages={2002314},
  year={2026},
  publisher={Springer}
}

@article{muller2023attending,
  title={Attending to graph transformers},
  author={M{\"u}ller, Luis and Galkin, Mikhail and Morris, Christopher and Ramp{\'a}{\v{s}}ek, Ladislav},
  journal={arXiv preprint arXiv:2302.04181},
  year={2023}
}

@article{vaswani2017attention,
  title={Attention is all you need},
  author={Vaswani, Ashish and Shazeer, Noam and Parmar, Niki and Uszkoreit, Jakob and Jones, Llion and Gomez, Aidan N and Kaiser, {\L}ukasz and Polosukhin, Illia},
  journal={Advances in neural information processing systems},
  volume={30},
  year={2017}
}

@article{hettige2024airphynet,
  title={Airphynet: Harnessing physics-guided neural networks for air quality prediction},
  author={Hettige, Kethmi Hirushini and Ji, Jiahao and Xiang, Shili and Long, Cheng and Cong, Gao and Wang, Jingyuan},
  journal={arXiv preprint arXiv:2402.03784},
  year={2024}
}

@article{xu2025sigrl,
  title={SIGRL: Sociologically-Informed Graph Representation Learning for Social Influence Prediction},
  author={Xu, Haowei and Gao, Chao and Li, Xianghua and Wang, Zhen},
  journal={IEEE Transactions on Network Science and Engineering},
  year={2025},
  publisher={IEEE}
}

@article{codina1998comparison,
  title={Comparison of some finite element methods for solving the diffusion-convection-reaction equation},
  author={Codina, Ramon},
  journal={Computer methods in applied mechanics and engineering},
  volume={156},
  number={1-4},
  pages={185--210},
  year={1998},
  publisher={Elsevier}
}

@article{bird2002transport,
  title={Transport phenomena},
  author={Bird, R Byron},
  journal={Applied Mechanics Reviews},
  volume={55},
  number={1},
  pages={R1--R4},
  year={2002},
  publisher={American Society of Mechanical Engineers Digital Collection}
}

@inproceedings{choi2023gread,
  title={Gread: Graph neural reaction-diffusion networks},
  author={Choi, Jeongwhan and Hong, Seoyoung and Park, Noseong and Cho, Sung-Bae},
  booktitle={International conference on machine learning},
  pages={5722--5747},
  year={2023},
  organization={PMLR}
}

@article{chen2018neural,
  title={Neural ordinary differential equations},
  author={Chen, Ricky TQ and Rubanova, Yulia and Bettencourt, Jesse and Duvenaud, David K},
  journal={Advances in neural information processing systems},
  volume={31},
  year={2018}
}

@article{dormand1980family,
  title={A family of embedded Runge-Kutta formulae},
  author={Dormand, John R and Prince, Peter J},
  journal={Journal of computational and applied mathematics},
  volume={6},
  number={1},
  pages={19--26},
  year={1980},
  publisher={Elsevier}
}

@article{bronstein2017geometric,
  title={Geometric deep learning: going beyond euclidean data},
  author={Bronstein, Michael M and Bruna, Joan and LeCun, Yann and Szlam, Arthur and Vandergheynst, Pierre},
  journal={IEEE Signal Processing Magazine},
  volume={34},
  number={4},
  pages={18--42},
  year={2017},
  publisher={IEEE}
}

@article{allen1979microscopic,
  title={A microscopic theory for antiphase boundary motion and its application to antiphase domain coarsening},
  author={Allen, Samuel M and Cahn, John W},
  journal={Acta metallurgica},
  volume={27},
  number={6},
  pages={1085--1095},
  year={1979},
  publisher={Elsevier}
}

@article{wang2022acmp,
  title={Acmp: Allen-cahn message passing for graph neural networks with particle phase transition},
  author={Wang, Yuelin and Yi, Kai and Liu, Xinliang and Wang, Yu Guang and Jin, Shi},
  journal={arXiv preprint arXiv:2206.05437},
  year={2022}
}

@article{barabasi1999emergence,
  title={Emergence of scaling in random networks},
  author={Barab{\'a}si, Albert-L{\'a}szl{\'o} and Albert, R{\'e}ka},
  journal={science},
  volume={286},
  number={5439},
  pages={509--512},
  year={1999},
  publisher={American Association for the Advancement of Science}
}

@article{karniadakis2021physics,
  title={Physics-informed machine learning},
  author={Karniadakis, George Em and Kevrekidis, Ioannis G and Lu, Lu and Perdikaris, Paris and Wang, Sifan and Yang, Liu},
  journal={Nature Reviews Physics},
  volume={3},
  number={6},
  pages={422--440},
  year={2021},
  publisher={Nature Publishing Group UK London}
}

@article{wang2021understanding,
  title={Understanding and mitigating gradient flow pathologies in physics-informed neural networks},
  author={Wang, Sifan and Teng, Yujun and Perdikaris, Paris},
  journal={SIAM Journal on Scientific Computing},
  volume={43},
  number={5},
  pages={A3055--A3081},
  year={2021},
  publisher={SIAM}
}

@article{de1986reaction,
  title={Reaction-diffusion equations for interacting particle systems},
  author={De Masi, Anna and Ferrari, Pablo A and Lebowitz, Joel L},
  journal={Journal of statistical physics},
  volume={44},
  number={3},
  pages={589--644},
  year={1986},
  publisher={Springer}
}

@incollection{chapman2015advection,
  title={Advection on graphs},
  author={Chapman, Airlie},
  booktitle={Semi-autonomous networks: Effective control of networked systems through protocols, design, and modeling},
  pages={3--16},
  year={2015},
  publisher={Springer}
}

@inproceedings{li2024predicting,
  title={Predicting long-term dynamics of complex networks via identifying skeleton in hyperbolic space},
  author={Li, Ruikun and Wang, Huandong and Piao, Jinghua and Liao, Qingmin and Li, Yong},
  booktitle={Proceedings of the 30th ACM SIGKDD Conference on Knowledge Discovery and Data Mining},
  pages={1655--1666},
  year={2024}
}

@article{bilovs2021neural,
  title={Neural flows: Efficient alternative to neural ODEs},
  author={Bilo{\v{s}}, Marin and Sommer, Johanna and Rangapuram, Syama Sundar and Januschowski, Tim and G{\"u}nnemann, Stephan},
  journal={Advances in neural information processing systems},
  volume={34},
  pages={21325--21337},
  year={2021}
}

@article{verma2024climode,
  title={Climode: Climate and weather forecasting with physics-informed neural odes},
  author={Verma, Yogesh and Heinonen, Markus and Garg, Vikas},
  journal={arXiv preprint arXiv:2404.10024},
  year={2024}
}

@article{wu2023advective,
  title={Advective diffusion transformers for topological generalization in graph learning},
  author={Wu, Qitian and Yang, Chenxiao and Zeng, Kaipeng and Nie, Fan and Bronstein, Michael and Yan, Junchi},
  journal={arXiv preprint arXiv:2310.06417},
  year={2023}
}

\appendix

\section{Theoretical Analysis}

\subsection{Analysis of Classical Opinion Dynamics}
\label{appa1}
In this section, we establish the theoretical basis for decoupling opinion dynamics into local, global, and endogenous levels by revisiting three most representative opinion dynamics models.

\noindent \textbf{DeGroot}~\cite{degroot1974reaching}:
Inspired by homogenization in sociology, DeGroot posits that the individual opinions evolve over discrete time steps through local interactions.
Given the context of real-world social platforms, the interaction of user $i$ is constrained to immediate neighbors $\mathcal{N}(i)$ within social network.
Specifically, the opinion of user $i$ updates according to a uniform averaging:
\begin{equation}
\label{degroot}
    x_i(t+1) = \frac{1}{|\mathcal{N}(i)|+1} x_i(t) + \frac{1}{|\mathcal{N}(i)|+1} \sum_{j \in \mathcal{N}(i)}  x_j(t),
\end{equation}
where $x_i(t)$ denotes the opinion value of user $i$ at time $t$. 
This formulation captures the mechanism of local consensus—the tendency of individuals to align their opinions with their local social circle.

\noindent \textbf{Hegselmann-Krause (HK) model}~\cite{hegselmann2015opinion}: 
Different from local influences, HK model introduces the principle of bounded confidence to describe global interactions.
It posits that individuals interact with any peer across the entire population $\mathcal{V}$ whose opinion is within a proximity threshold $\epsilon$ beyond physical network ties:
\begin{equation}
\label{hk}
x_i(t+1) = \frac{1}{|\mathcal{N}_i^{\epsilon}(t)|} \sum_{j \in \mathcal{N}_i^{\epsilon}(t)} x_j(t),
\end{equation}
where $\mathcal{N}_i(t, \epsilon) = \{j \in \mathcal{V} / \{i\} \mid |x_i(t) - x_j(t)| \le \epsilon\}$ denotes the set of selected neighbors of user $i$ at time $t$.
Through this threshold-based filtering, the HK model captures how selective assimilation leads to population-wide opinion fragmentation and clustering.

\noindent \textbf{Friedkin-Johnsen (FJ) model}~\cite{friedkin1990social}:
The FJ model extends DeGroot model by incorporating "stubbornness" to describe the endogenous reaction of individuals. It represents opinion evolution as a trade-off between social influence and an internal cognitive anchor:
\begin{equation}
\label{fjmodel}
x_i(t+1) =  \alpha x_i(0) + \frac{1}{|\mathcal{N}(i)|} \sum_{j \in \mathcal{N}(i)}  x_j(t),
\end{equation}
where $x_i(0)$ represents the initial opinion, and $\alpha$ denotes the susceptibility coefficient to social influence.
Through this internal-external interplay, FJ model captures the endogenous persistence of disagreement despite social pressure.

Notably, we apply the uniform aggregation in the aforementioned models for notational simplicity. 
Synthesizing these classic opinion models, we observe that they each emphasize distinct dimensions of dynamics: local, global, and endogenous levels, while neglecting the intricate interplay among them.
Meanwhile, the patterns across these three tiers align perfectly with the inherent logic of interacting particle theory~\cite{de1986reaction}, thereby providing a theoretical foundation for the introduction of the DCR system.

\subsection{Analysis of Physic-Informed Modeling}
\label{appa2}
This section present detailed theoretical derivations to elucidate the connections between the DCR system components with the DeGroot, HK, FJ models, aiming to bridge the proposed physical perspective with established opinion dynamics models.

\noindent \textbf{Diffusion.}
For the continuous dynamics of diffusion in Equation~\ref{dif}, we apply the first-order forward Euler method:
\begin{equation}
    \frac{x_i(t + \Delta t) - x_i(t)}{\Delta t} = \sum_{j \in \mathcal{N}(i)} w_{ij}(x_j(t) - x_i(t)).
\end{equation}
Rearranging the terms for the opinion state at the next time step $t + \Delta t$, we obtain:
\begin{equation}
x_i(t + \Delta t) = x_i(t) + \Delta t \sum_{j \in \mathcal{N}(i)} w_{ij} \big( x_j(t) - x_i(t) \big).
\end{equation}
We normalize the time step by setting $\Delta t = 1$, and this yields the discrete update rule:
\begin{equation}
\begin{aligned}
x_i(t+1) &= x_i(t) + \sum_{j \in \mathcal{N}(i)} w_{ij} \big( x_j(t) - x_i(t) \big) \\
         &= x_i(t) - \sum_{j \in \mathcal{N}(i)} w_{ij} x_i(t) + \sum_{j \in \mathcal{N}(i)} w_{ij} x_j(t) \\
         &= \big(1 - \sum_{j \in \mathcal{N}(i)} w_{ij} \big) x_i(t) + \sum_{j \in \mathcal{N}(i)} w_{ij} x_j(t).
\end{aligned}
\end{equation}
Under the assumption of uniform diffusion, i.e., $w_{ij} = \frac{1}{|\mathcal{N}(i)|+1}$, the diffusion proecess simplifies to: 
\begin{equation}
\begin{aligned}
x_i(t+1) &= \big(1 - \sum_{j \in \mathcal{N}(i)} \frac{1}{|\mathcal{N}(i)|+1} \big) x_i(t) + \sum_{j \in \mathcal{N}(i)}\frac{1}{|\mathcal{N}(i)|+1} x_j(t)\\
         &= \left(1 - \frac{|\mathcal{N}(i)|}{|\mathcal{N}(i)|+1} \right) x_i(t) + \frac{1}{|\mathcal{N}(i)|+1} \sum_{j \in \mathcal{N}(i)}  x_j(t),
\end{aligned}
\end{equation}
which is equal to Equation~\ref{degroot} and positions that diffusion is a generalization of DeGroot model to capture local assimilation patterns.

\noindent \textbf{Convection.}
Similar to the diffusion, we adopt the first-order forward Euler method and set $\Delta t = 1$ to Equation~\ref{con}:
\begin{equation}
\begin{aligned}
   x_i(t+1) &= x_i(t) + \sum_{\forall j | j\rightarrow i} \vec{w}_{ji} x_j(t)  - \sum_{\forall k | i \rightarrow k} \vec{w}_{ik} x_i(t) \\
            &= \big(1 - \sum_{\forall k | i \rightarrow k} \vec{w}_{ik} \big) x_i(t) + \sum_{\forall j | j\rightarrow i} \vec{w}_{ji} x_j(t).
\end{aligned}
\end{equation}
By simplifying the global directed transport into undirected propagation, we obtain:
\begin{equation}
    x_i(t+1) = \big(1 - \sum_{j \in \mathcal{V} / \{i\}} w_{ij} \big) x_i(t) + \sum_{j \in \mathcal{V} / \{i\}} w_{ij} x_j(t).
\end{equation}
Treating global users as interacting parties and representing them in set form, i.e., $\mathcal{N}_i(t) = \mathcal{V} / \{i\}$. Under the assumption of uniform aggregation, we can obtain:
\begin{equation}
\begin{aligned}
    x_i(t+1) &= \big(1 - \sum_{j \in \mathcal{N}_i(t)} \frac{1}{|\mathcal{N}_i(t)|} \big) x_i(t) + \sum_{j \in \mathcal{N}_i(t)} \frac{1}{|\mathcal{N}_i(t)|} x_j(t) \\
    &= \frac{1}{|\mathcal{N}_i(t)|} \sum_{j \in \mathcal{N}_i(t)}  x_j(t).
\end{aligned}
\end{equation}
By introducing the threshold $\epsilon$ to further constrain $\mathcal{N}_i(t)$, the convection process reduces to the HK model as illustrated in Equation~\ref{hk}.
Consequently, the bounded confidence models in sociology can be viewed as special cases of the convection process.

\noindent \textbf{Reaction.}
Regarded as a non-conservative component, the reaction disrupts the assumption of a closed system and is often integrated with diffusion and convection to characterize open-system dynamics.
Here, we consider a basic configuration: a reaction-diffusion process where the a initial state $x_i(0)$ serves as the reaction term. 
The resulting update process for user opinions is given by:
\begin{equation}
    x_i(t+1) = \big(1 - \sum_{j \in \mathcal{N}(i)} w_{ij} \big) x_i(t) + \sum_{j \in \mathcal{N}(i)} w_{ij} x_j(t) + \delta x_i(0),
\end{equation}
where $\delta$ is the weighting factor to balance the interplay between diffusion and reaction.
By assuming uniform aggregation of neighbors with $w_{ij} = \frac{1}{|\mathcal{N}(i)|}$ as before, we can obtain:
\begin{equation}
    \begin{aligned}
    x_i(t+1) &= \big(1 - \sum_{j \in \mathcal{N}(i)} \frac{1}{|\mathcal{N}(i)|} \big) x_i(t) + \sum_{j \in \mathcal{N}(i)} \frac{1}{|\mathcal{N}(i)|} x_j(t) + \delta x_i(0) \\
    & =  \delta x_i(0) + \frac{1}{|\mathcal{N}(i)|} \sum_{j \in \mathcal{N}(i)} x_j(t),
    \end{aligned}
\end{equation}
which is equivalent to the FJ model in Equation~\ref{fjmodel}.
Here, the reaction term serves to reinforce the user's initial belief, analogous to the concept of stubbornness in social psychology.

\section{Dataset Details.}

\subsection{Real-world Datasets}
\label{appdata1}
\noindent \textbf{Dataset sources.}
Regarding the data sources, the Delhi Election dataset originates from~\cite{de2019learning}, whereas the other three datasets are drawn from~\cite{li2025unigo}. 
In terms of social platforms, the COVID-19 dataset is collected from Weibo, whereas the others are retrieved from X (Twitter).
User opinions are defined according to the nature of each dataset. In political contexts (Delhi and U.S. Elections), opinions represent partisan leaning. 
For Israel-Palestine, they reflect the acceptance or rejection of conflict-related rumors, while in COVID-19, they characterize the prevailing sentiment ranging from optimism to pessimism regarding the global health crisis.

\noindent \textbf{Network construction.}
To filter out irrelevant information, corporate accounts and social bots are removed from all four real-world datasets. 
The U.S. Election, Israel-Palestine, and COVID-19 datasets is restricted to active users, with interaction thresholds set at 50, 8, and 100, respectively.
The social network is static and constructed by establishing edges through retweet or comment relationships, followed by the retention of the largest connected component~\cite{li2025unigo}.

\noindent \textbf{Opinion annotation.}
User opinions are operationalized as quantitative representations based on user textual posts to specific topics, employing a continuous scale from -1 to 1 that maps negative to positive sentiments, aligning with established opinion dynamics frameworks.
Opinion value is annotated using LLM, complemented by human verification to ensure data quality. A structured prompt engineering approach is adopted for GPT-3.5, utilizing context-driven reasoning chains formulated as follows: event context, opinion detection, step-by-step think through chain of thought (COT).

\noindent \textbf{Data preprocessing.}
To prepare the datasets for analysis, we first impute missing entries and then perform point-wise linear interpolation to enhance temporal resolution.
Specifically, the number of interpolated points is set to 2 for the Delhi Election, 3 for the U.S. Election, 21 for the Israel-Palestine, and 6 for the COVID-19.

\noindent \textbf{Quantitative analysis.}
Quantitative analysis of opinion dynamics reveals distinct temporal and structural characteristics across the real-world datasets.
As summarized in Table~\ref{appopinion_stats}, we introduce two metrics: Mean Transition Steps, defined as the average step for all users to transition their opinions before data interpolation, and Mean Absolute Shifts, which quantifies the average absolute difference between the initial and terminal opinion states.
These two metrics provide insights into the varying intensity and duration of opinion stabilization across different social contexts.
\begin{table}[htbp]
  \centering
  \caption{Quantitative characteristics of real-world datasets.}
  \label{appopinion_stats}
  \resizebox{0.4\textwidth}{!}{
  \begin{tabular}{l cc}
    \toprule
    \textbf{Dataset} & {\textbf{Mean Transition Steps}} & {\textbf{Mean Absolute Shift}} \\
    \midrule
    Delhi Election & 5.1243 & 0.1548 \\
    U.S. Election  & 7.6356 & 0.5193 \\
    Israel-Palestine  & 3.3149 & 0.6043 \\
    COVID-19       & 2.9048 & 0.2736 \\
    \bottomrule
  \end{tabular}
  }
\end{table}
\begin{figure*}[htbp]
    \centering
    \begin{subfigure}[b]{0.32\linewidth}
        \centering
        \includegraphics[width=\linewidth]{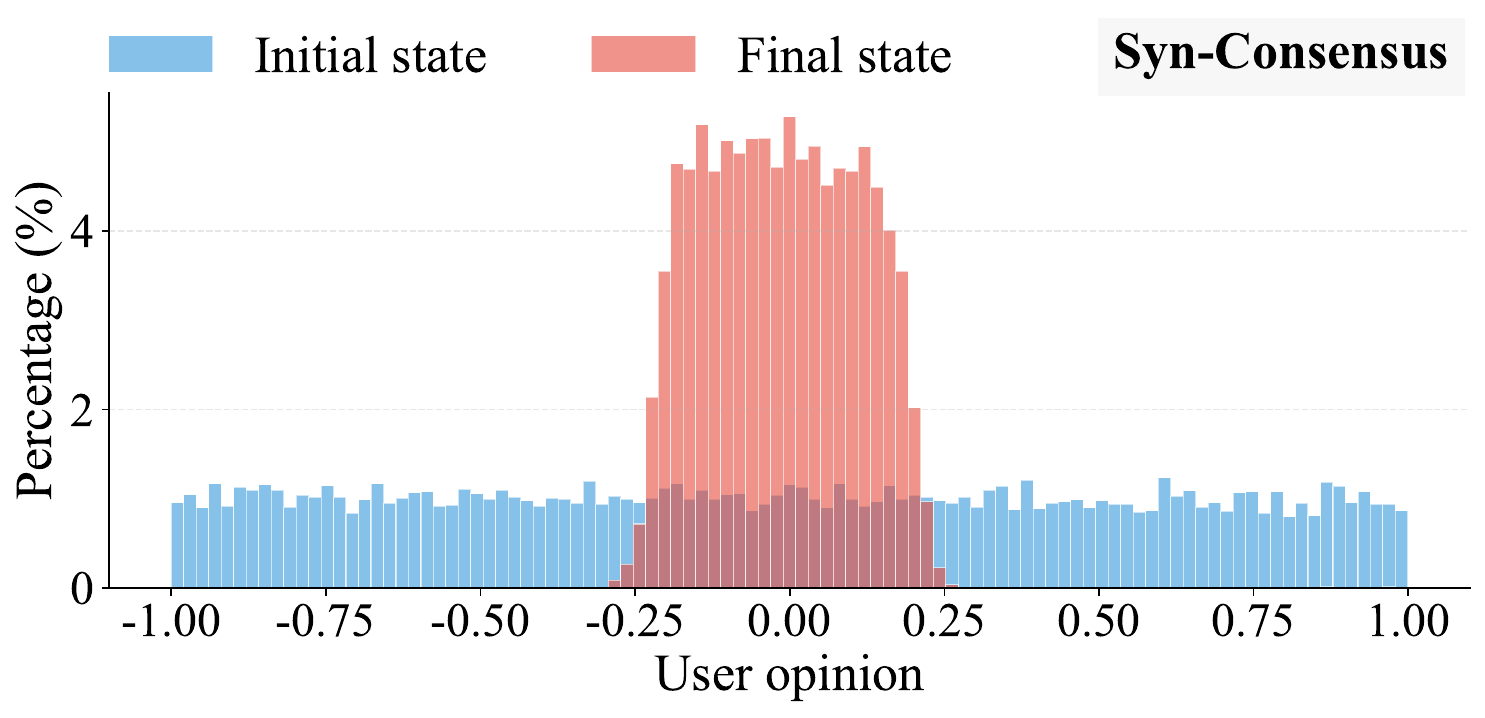}
    \end{subfigure}
    \begin{subfigure}[b]{0.32\linewidth}
        \centering
        \includegraphics[width=\linewidth]{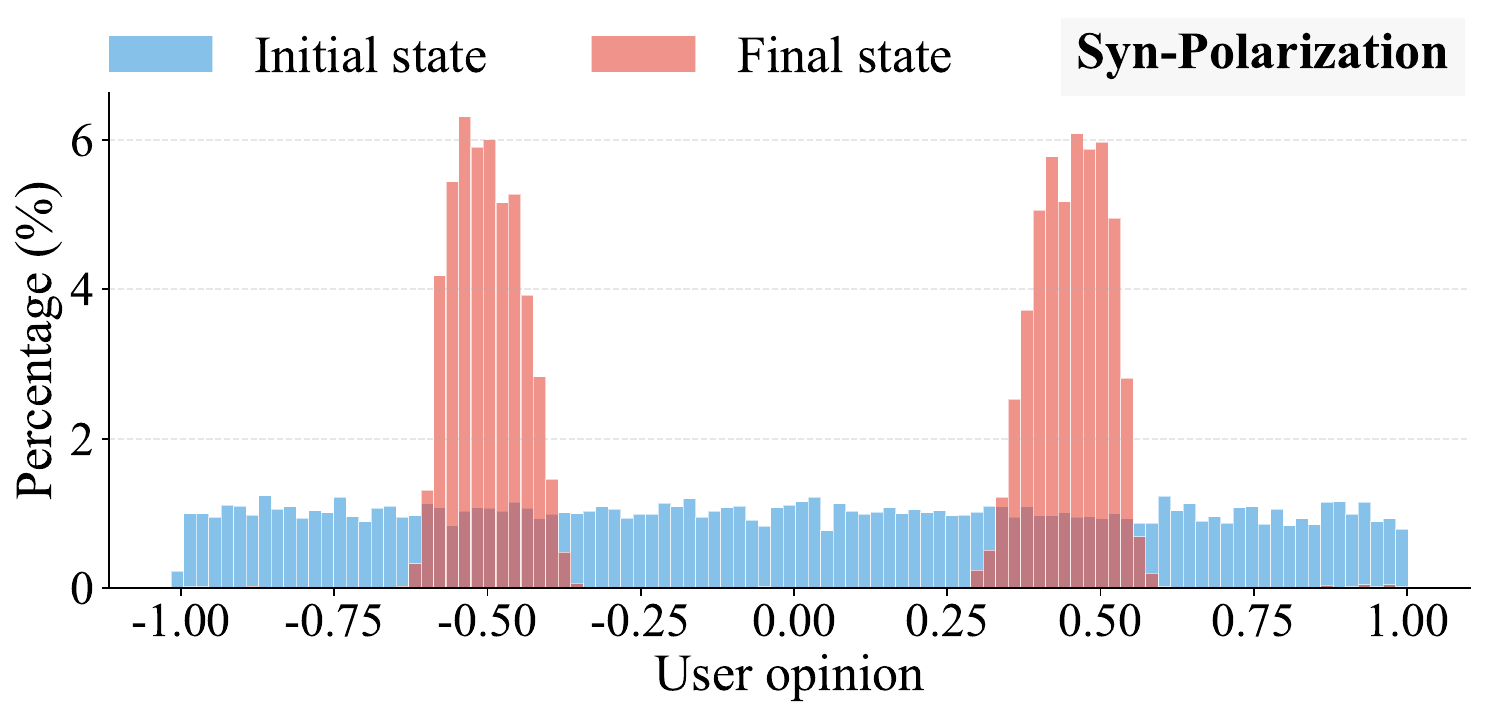}
    \end{subfigure}
    \begin{subfigure}[b]{0.32\linewidth}
        \centering
        \includegraphics[width=\linewidth]{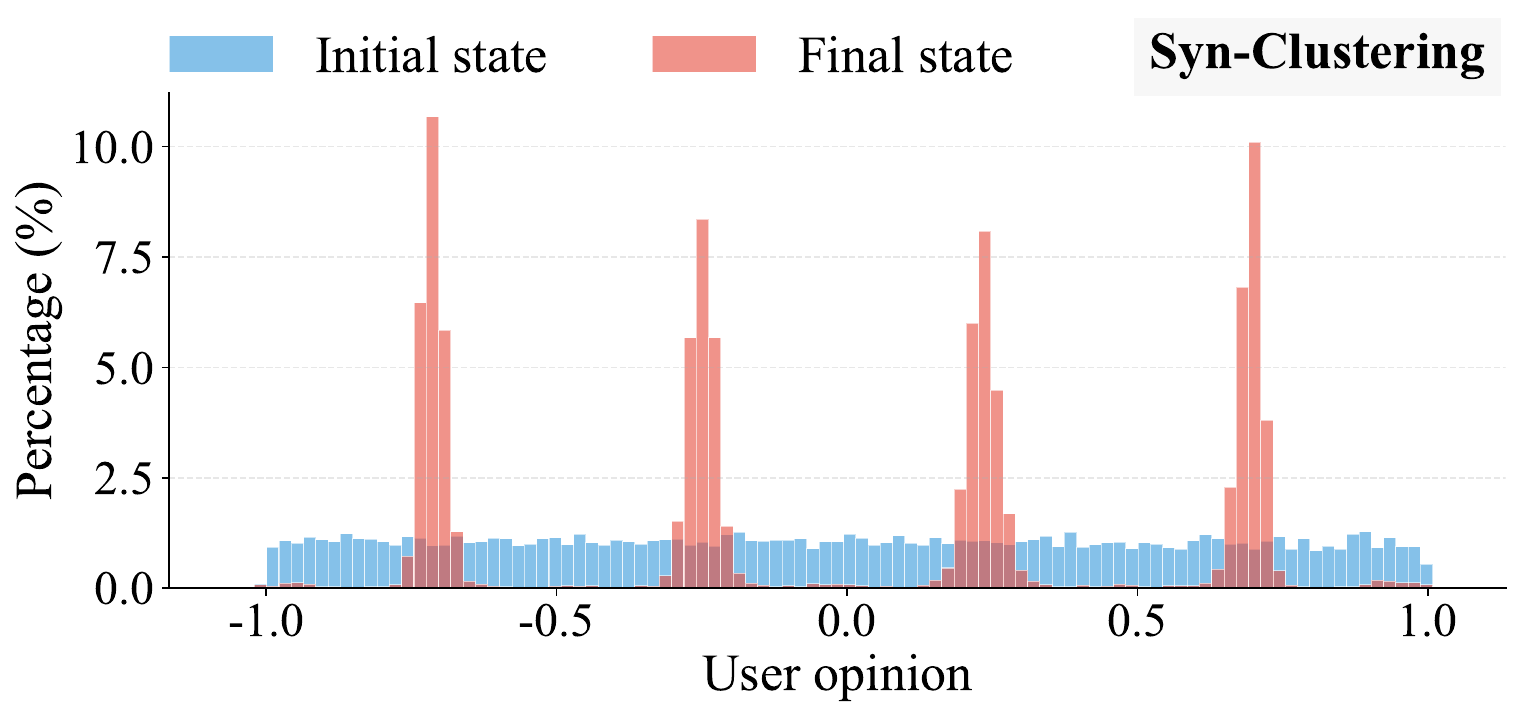}
    \end{subfigure}
    \caption{Visualization of user opinion distribution in synthetic datasets.} 
    \label{syndatadis}
\end{figure*}

\subsection{Synthesis Datasets}
\label{appdata2}
To supplement the real-world datasets, we additionally generate synthetic datasets under typical patterns: Consensus, Polarization, and Clustering.
The simulation involves 10,000 users whose initial opinions are uniformly distributed,
and social structure is built upon a BA scale-free graph.
We follow the update rule in~\cite{li2025unigo} to perform opinion simulation:
\begin{equation}
    x_i(t+1) = \lambda x_i(0) +  (1-\lambda) \sum_{j\in \mathcal{N}_i^{\epsilon}(t)} \frac{1}{|\mathcal{N}_i^{\epsilon}(t)|}x_j(t)+ \eta \xi_i(t),
\end{equation}
where $\lambda$ is the stubbornness coefficient to balance initial states and external drivers, $\epsilon$ denotes the bounded confidence threshold to user neighbors for modulating opinion clustering, $\xi_i(t) \sim \mathcal{N}(0,1)$ is a standard Gaussian distribution and $\eta$ represents the noise level.
It is worth noting that $\epsilon$ is closely associated with opinion evolution patterns. Specifically, a smaller value of  $\epsilon$ leads to a higher degree of opinion fragmentation trends (i.e., more clusters), and vice versa.
The parameter configurations for synthetic datasets under the three specific modes are presented in Table~\ref{appsyndata}.
To avoid reaching opinion convergence, we extract the first 50 time steps and apply linear interpolation to generate 400 temporal points. 
The initial and final distributions of all user opinions are illustrated in Figure~\ref{syndatadis}.
\begin{table}[ht]
    \setlength{\tabcolsep}{4pt}
    \centering
    \caption{The parameter configurations for synthetic datasets. }
    \label{appsyndata}
    \resizebox{0.47\textwidth}{!}{
    \begin{tabular}{c c c c c c c c}
    \toprule
    \textbf{Dataset} & \textbf{\# Nodes} & \textbf{Network} & \textbf{\# Edges} & \textbf{Time steps} & \textbf{$\lambda$} & \textbf{$\epsilon$} & \textbf{$\eta$}\\
    \midrule
    Syn-Consensus & 10,000 & BA-Graph & 99,000 & 400 & 0.2 & 0.5 & 0.015 \\
    Syn-Polarization & 10,000 & BA-Graph & 99,000 & 400 & 0.1 & 0.3 & 0.015 \\
    Syn-Clustering & 10,000 & BA-Graph & 99,000 & 400 & 0.15 & 0.2 & 0.015 \\
    \bottomrule
    \end{tabular}
    }
\end{table}


\section{Implementation Details}
\label{appimp}
\noindent \textbf{Model implementation.}
In the default settings, we adopt a one-layer GRU as the encoder and a two-layer MLP as the decoder.
We set the input context of encoder as 30 T, and view it as one time step for the system state.
We employ the Adam optimizer with a weight decay of 5e-5, fixing the training duration at 200 epochs. 
We identify the optimal hidden dimension, learning rate, and batch size via hyperparameter tuning.
For neural dynamics module, we implement the diffusion via a single GCN layer and convection through one described attention layer, respectively. 
Specifically, we employ ReLU as the activation function for both the diffusion and convection terms.
For the reaction term, we report the results of the best-performing one among the candidate source, linear, and non-linear terms.
For the gating weights within neural dynamics module, we consistently initialize the $\omega$ and $\delta$ to 0.5.
For the ODE solver, we adopt the RK-4 numerical integration method, simply set the integration interval as $t=1.0$ for one system step.

\noindent \textbf{Hyperparameter settings.}
We conduct a grid search strategy for key hyperparameters to achieve optimal model performance.
The following list details the search space:
\begin{itemize}
    \item learning rate within \{ 0.001, 0.005, 0.01, 0.05 \}
    \item hidden dimension within \{ 8, 16, 32, 64, 128 \}
    \item batch size within \{ 1, 4, 16, 32, 64 \}
\end{itemize}
\begin{figure*}[htbp]
    \centering 
        \includegraphics[width=0.79\textwidth]{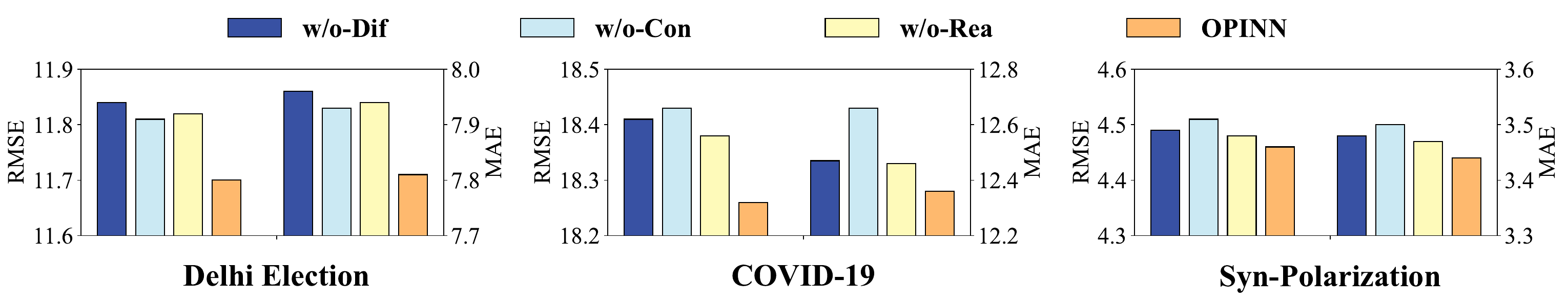} 
    \caption{The additional ablation results of opinion dynamics mechanisms ($\times 10^{-2}$).} 
    \label{appabpro} 
\end{figure*}
\section{Further Details of Baselines}
\subsection{Baseline Introduction}
\label{appbaseline1}
The baseline models in this paper are described as follows:
\begin{itemize}

\item \textbf{Voter}~\cite{yildiz2010voting}: 
The Voter model operates by letting each user randomly select another neighboring user at each time step and adopt that user’s opinion.

\item \textbf{DeGroot}~\cite{degroot1974reaching}: 
The DeGroot model simulates opinion dynamics by having user update their opinion as a weighted average of their connected neighbors' opinions at each time step.
In our implementation, the connections between users are defined by the given adjacency matrix.

\item \textbf{FJ (Friedkin–Johnsen model)}~\cite{friedkin1990social}: 
The FJ model extends the DeGroot model with a stubbornness coefficient to model scenarios where individuals retain initial opinions.

\item \textbf{HK (Hegselmann-Krause model)}~\cite{hegselmann2015opinion}:    
The HK model introduces confidence thresholds to simulate selective interaction, whereby individuals only interact with peers whose opinions fall within a prescribed range.

\item  \textbf{GCN}~\cite{kipf2016semi}:  
GCN is a representative graph neural network that aggregates neighbor information using fixed spectral filters.
Following the configuration in~\cite{li2025unigo}, we utilize the opinion values within the input window for users as node features to predict future opinion trends.

\item  \textbf{GAT}~\cite{velivckovic2017graph}:
GAT is also a classic graph neural network that captures structural dependencies via attention mechanism. 

\item \textbf{DeepInf}~\cite{qiu2018deepinf}: DeepInf is a typical GNN-based framework for social influence prediction. In terms of implementation, we leverage the given adjacency matrix for user connections, build user embeddings through a linear layer, and adopt the GAT~\cite{velivckovic2017graph} for structural encoding, in accordance with the task settings established in~\cite{li2025unigo}.

\item \textbf{iTransformer}~\cite{liu2023itransformer}:
iTransformer is a advanced time-series forecasting model based on the Transformer architecture. Under our task setting, we treat the observable opinions of each user as a token and compute global attention across users to aggregate information for future trends prediction.

\item \textbf{SGFormer}~\cite{wu2023sgformer}: SGFormer is a representative large-graph transformer that integrates global attention with GNN-captured local patterns. In our setting, we utilize the
input-window opinions for users as node features and adopt the GAT~\cite{velivckovic2017graph} as the local components.

\item \textbf{UniGO}~\cite{li2025unigo}:
UniGO is a specialized GNN-based framework tailored for modeling opinion dynamics within social networks.
This framework leverages the graph coarsen-refine method to uncover community patterns on graphs.
In our implementation, we use a two-alyer GCN~\cite{kipf2016semi} as the encoder to obtain the graph skeleton and adopt a single linear layer for the graph refinement module. 

\item  \textbf{SINN}~\cite{okawa2022predicting}:
SINN draws on the concept of PINNs to introduce a Sociologically-Informed Neural Network framework.
This framework formalizes classic models of sociological opinion dynamics into ordinary differential equations, integrating them as penalty terms to regularize the training process of neural network.
For the neural module, we employ a three-layer Feedforward Neural Network (FNN) to encoder user opinions, and report the results obtained using the best-performing sociological prior.

\item  \textbf{SIGNN}~\cite{yang2025sociologically}:
Similar to SINN, SIGNN employs the stochastic bounded confidence model as sociological priors to constrain neural networks. Additionally, the model employs GCN~\cite{kipf2016semi} to encode user opinions and incorporates a time-mixing module to aggregate historical opinions.

\item \textbf{ODENet}~\cite{lv2023unified}:
ODENet is a opinion dynamics-inspired neural message passing framework, which reveals the explicit connection between opinion dynamics and neural message passing.
Inspired by the dynamics of HK opinion model, this method optimizes the messaging passing process by defining adaptive aggregation thresholds, thereby facilitating selective feature aggregation.

\item \textbf{GRAND}~\cite{chamberlain2021grand}:
GRAND is a classic neural message passing framework inspired by physical diffusion process.
In our implementation, we employ a two-layer MLP to project user opinions into the latent space, while utilizing graph attention to model the neural diffusion process in the spirit of Neural ODEs.
The model is trained to output the users' subsequent opinion values.

\item \textbf{GREAD}~\cite{choi2023gread}:
Building upon GRAND, GREAD extends the neural diffusion process into a more expressive diffusion-reaction process. 
In our experiments, the configuration of the reaction term remains consistent with \textsc{Opinn}.

\item \textbf{AdvDifformer}~\cite{wu2023advective}: Inspired by the physical diffusion-advection process, AdvDifformer integrates graph neural network with Transformer to extend the solitary diffusion process, achieving a unification of local and global patterns.

\end{itemize}

\subsection{Baseline Analysis}
\label{appbaseline2}
In this section, we review and analyze all the baselines from the proposed physics-inspired system perspective.
We categorize current baselines based on the three physical components within the DCR system: diffusion, convection, and reaction, as summarized in Table~\ref{bslana}. 
The superiority of \textsc{Opinn} stems from its ability to model diverse opinion evolution patterns, grounded in a broad spectrum of physical phenomena~\cite{choi2023gread,wu2023advective}.

\begin{table}[htbp]
  \centering
  \caption{Analysis of baselines based on DCR components.}
  \label{bslana}
  \resizebox{0.44\textwidth}{!}{
  \begin{tabular}{ccccccc} 
    \toprule
    \textbf{Models} & \textbf{Phyisic-informed} & \textbf{Data-driven} & \textbf{Diffusion} & \textbf{Convection} & \textbf{Reaction} \\
    \midrule
    Voter~\cite{yildiz2010voting} & - & - & \checkmark & - & - \\
    DeGroot~\cite{degroot1974reaching} & - & - & \checkmark & - & - \\
    FJ~\cite{friedkin1990social} & - & - & \checkmark & - & \checkmark \\
    HK~\cite{hegselmann2015opinion} & - & - & - & \checkmark & -  \\
    \midrule
    GCN~\cite{kipf2016semi} & - & \checkmark & \checkmark & - & -\\
    GAT~\cite{velivckovic2017graph} & - & \checkmark & \checkmark & - & -\\
    DeepInf~\cite{qiu2018deepinf} & - & \checkmark & \checkmark & - &  -\\
    iTransformer~\cite{liu2023itransformer} & - & \checkmark & - & \checkmark & - \\
    SGFormer~\cite{wu2023sgformer} & - & \checkmark & \checkmark & \checkmark & -  \\
    UniGO~\cite{li2025unigo} & - & \checkmark & \checkmark & \checkmark & - \\
    \midrule
    SINN~\cite{okawa2022predicting} & \checkmark & \checkmark & - & \checkmark & - \\
    SIGNN~\cite{yang2025sociologically} & \checkmark & \checkmark & - & \checkmark & - \\
    ODENet~\cite{lv2023unified} & \checkmark & \checkmark & \checkmark & - & - \\
    GRAND~\cite{chamberlain2021grand} & \checkmark & \checkmark & \checkmark & - & - \\
    GREAD~\cite{choi2023gread} & \checkmark & \checkmark & \checkmark & - & \checkmark \\
    AdvDifformer~\cite{wu2023advective} & \checkmark & \checkmark & \checkmark & \checkmark & -\\
    \midrule
    \textsc{Opinn} (Ours) & \checkmark & \checkmark & \checkmark & \checkmark & \checkmark  \\
    \bottomrule
  \end{tabular}
  }
\end{table}

\section{Supplementary Results}

\subsection{Opinion Dynamics Mechanisms}
\label{appodm}

Here, we provide the additional ablation results on opinion dynamics mechanisms for \textsc{Opinn} in Figure~\ref{appabpro}. 
The findings are consistent with those in the main text, further underscoring the importance of the complete physical process.
Further, we present the values of the learnable parameters $\omega$ and $\delta$ within neural dynamics module, as presented in Table~\ref{appgw}.
It is observed that diffusion coefficients are larger for Israel-Palestine and Election datasets. 
Therefore, removing the diffusion leads to a significant drop in model performance in ablation results.
Similarly, for COVID-19 and the synthetic dataset, the convection weights are larger; and removing the convection results in decreased performance.
Moreover, real-world datasets show larger reaction coefficients than synthetic ones, suggesting more complex endogenous user changes in practical scenarios.
The preceding observations offer insights into the mechanistic interpretability of our proposed model. 
Notably, we recognize the feasibility of customizing user-specific gating weights within neural opinion dynamics; however, to maintain computational overhead when scaling to large graphs and avoid the risk of overfitting~\cite{wang2022acmp} issue, we opt for a more parsimonious approach in this paper.

\begin{table}[htbp]
    \setlength{\tabcolsep}{4pt}
    \centering
    \caption{The presentation of gating weights within \textsc{Opinn}.}
    \label{appgw}
    \resizebox{0.44\textwidth}{!}{
    \begin{tabular}{c c c c }
    \toprule
    \textbf{Dataset} & \textbf{Diffusion weight ($\omega$)} & \textbf{Convection weight ($1-\omega$)} & \textbf{Reaction weight ($\delta$)} \\
    \midrule
    Delhi Election & 0.59 & 0.41 &  0.55\\ 
    U.S. Election  & 0.55 & 0.45 &  0.69\\
    Israel-Palestine & 0.64 & 0.36 & 0.63\\
    COVID-19 & 0.43 & 0.57 &  0.53\\
    Syn-Consensus &  0.46 & 0.54 & 0.29\\
    Syn-Polarization & 0.49 & 0.51 &  0.30  \\
    Syn-Clustering & 0.47 & 0.53 &  0.36\\
    \bottomrule
    \end{tabular}
    }
\end{table}
\begin{table}[htbp]
    \centering
    \caption{Additional results on the reaction term ($\times 10^{-2}$).}
    \small
    \label{appreaction}
    \resizebox{0.36\textwidth}{!}{
    \begin{tabular}{@{}c cc cc cc@{}} 
        \toprule
        \multirow{2}{*}{\textbf{Reaction term}} & \multicolumn{2}{c}{\textbf{Delhi Election}} & \multicolumn{2}{c}{\textbf{Israel-Palestine}} & \multicolumn{2}{c}{\textbf{Syn-Clustering}}  \\
        \cmidrule(lr){2-3} \cmidrule(lr){4-5} \cmidrule(lr){6-7} 
        &RMSE & MAE & RMSE & MAE & RMSE & MAE \\
        \midrule
        Source term & \underline{12.07} & \underline{8.05} & 26.12 & 18.97 & 4.74 & 3.44 \\
        Linear term & 12.26 & 8.33 &  \underline{25.97} &  \underline{17.94} & \underline{4.67} & \underline{3.39} \\
        Non-linear term & \textbf{11.70} & \textbf{7.81} & \textbf{24.76} & \textbf{16.32} & \textbf{4.60} & \textbf{3.35}  \\
        \bottomrule
    \end{tabular}
    }
\end{table}
\subsection{Reaction Term}
\label{apprt}
We provide the additional results on the reaction term for \textsc{Opinn} in Table~\ref{appreaction}.
Taking together results from the main body and appendix, it is observed that for synthetic datasets with distinct evolutionary patterns, linear reactions perform optimally on the
Syn-Consensus, whereas nonlinear reactions are superior for Syn-Polarization and Syn-Clustering. 
This stems from the fact that consensus evolution is essentially a physical dissipation process, which is inherently linear. 
Conversely, polarization and clustering correspond to phase separation and multi-stable dynamics in physics, both of which are quintessential nonlinear.
In our experiments, we report the best results among three reaction terms. However, we recommend nonlinear reaction when dealing with complex, real-world scenarios.

\subsection{Model Complexity}
\label{appmc}
\noindent \textbf{Computation overhead.}
We present the computational time for one training epoch and model size of \textsc{Opinn} as well as typical baselines on U.S. Election dataset in Table~\ref{appms}. 
Despite a slight increase in parameters compared to SINN and DeepInf, \textsc{Opinn} achieves superior performance with a relatively compact model size, thereby avoiding the computation consumption associated with overly complex architectures.
Considering the optimal performance, \textsc{Opinn} offers an acceptable trade-off between speed and performance, requiring 0.49s for training and 0.07s for inference.

\begin{table}[htbp]
    \setlength{\tabcolsep}{4pt}
    \centering
    \caption{The comparison of model size and computation time on the U.S. Election dataset.}
    \label{appms}
    \resizebox{0.43\textwidth}{!}{
    \begin{tabular}{c c c c c c}
    \toprule
    \textbf{Model} & \textbf{\textsc{Opinn}} & \textbf{iTransformer} & \textbf{SINN} & \textbf{DeepInf}&\textbf{AdvDifformer} \\
    \midrule
    Parameter number & 31,521 & 41,470 &  20,574 & 24,799 & 39,455\\ 
    Model size (MB)  & 0.12 & 0.14 & 0.08 & 0.09 & 0.15 \\
    Training time (s) &  0.49 & 0.64 & 0.22 & 0.29 & 0.35\\
    Inference time (s) & 0.07 & 0.11 & 0.03 & 0.05 & 0.06\\
    \bottomrule
    \end{tabular}
    }
\end{table}

\noindent \textbf{Complexity of convection.}
\label{appcc}
The convection involves a global attention computation, which is an operation with quadratic time complexity with respect to user number $N$, leading to the main computational bottleneck when generalizing to large-scale graphs.
Inspired by the attention linearization, we consider using it to implement the convection to improve computational efficiency. 
We implement the following two variants for comparison:
\textbf{Vanilla-Atten}: replace the proposed convection scheme via the vanilla attention~\cite{vaswani2017attention}.
\textbf{Linear-Atten}: replace the proposed convection scheme with the linear global attention in~\cite{wu2023sgformer}.
As shown in Table~\ref{convection}, we present the prediction performance of \textsc{Opinn} and two variants, along with the training time required for one epoch.
The proposed convection scheme consistently outperforms the variants, demonstrating the effectiveness 
of our convection design.
Although \textsc{Opinn}'s computation time is slightly longer than that required by linearized scheme, its consumption is competitive given that it guarantees optimal performance.
In most cases, the vanilla attention outperforms the linearized scheme, reflecting the trade-off between performance and computational efficiency.
On Israel-Palestine, vanilla scheme underperforms the linear scheme, likely due to the attention over-dispersion phenomenon caused by the excessive number of nodes. 
However, our convection design remains unaffected by this issue.
Furthermore, subgraph sampling and active user filtering also serve as effective strategies for addressing the computational bottleneck when processing expansive social networks.

\begin{table}[htbp]
    \centering
    \caption{The comparison of model performance and computation time for \textsc{Opinn} variants on real-world datasets.}
    \small
    \label{convection}
    \resizebox{0.49\textwidth}{!}{
    \begin{tabular}{@{}c ccc ccc ccc ccc@{}} 
        \toprule
        \multirow{2}{*}{\textbf{Model}} & \multicolumn{3}{c}{\textbf{Delhi Election}} & \multicolumn{3}{c}{\textbf{U.S. Election}} & \multicolumn{3}{c}{\textbf{Israel-Palestine}} & \multicolumn{3}{c}{\textbf{COVID-19}} \\
        \cmidrule(lr){2-4} \cmidrule(lr){5-7} \cmidrule(lr){8-10} \cmidrule(lr){11-13}
        & RMSE & MAE & Time (s) & RMSE & MAE & Time (s) & RMSE & MAE & Time (s) & RMSE & MAE & Time (s)\\
        \midrule
        Vanilla-Atten & 12.08 & 8.11 & 0.67 & 46.01 & 29.50 & 0.42 & 28.85 & 19.77 & 24.89 & 19.08 & 13.23 & 0.45\\
        Linear-Atten & 12.46 & 8.59 & 0.42 & 47.18 & 30.91 & 0.26 & 26.87 & 18.04 & 17.53 & 19.34 & 13.51 & 0.29\\
        \textsc{Opinn} (Ours) & 11.70 & 7.81 & 0.71 & 44.43 & 27.29 & 0.49 & 24.76 & 16.32 & 31.09 & 18.26 & 12.36 & 0.52 \\
        \bottomrule
    \end{tabular}
    }
\end{table}

\subsection{Case Study}
\label{appcs}
\begin{figure}[htbp]
    \centering
    \begin{subfigure}{0.49\textwidth}
        \centering
        \includegraphics[width=\textwidth]{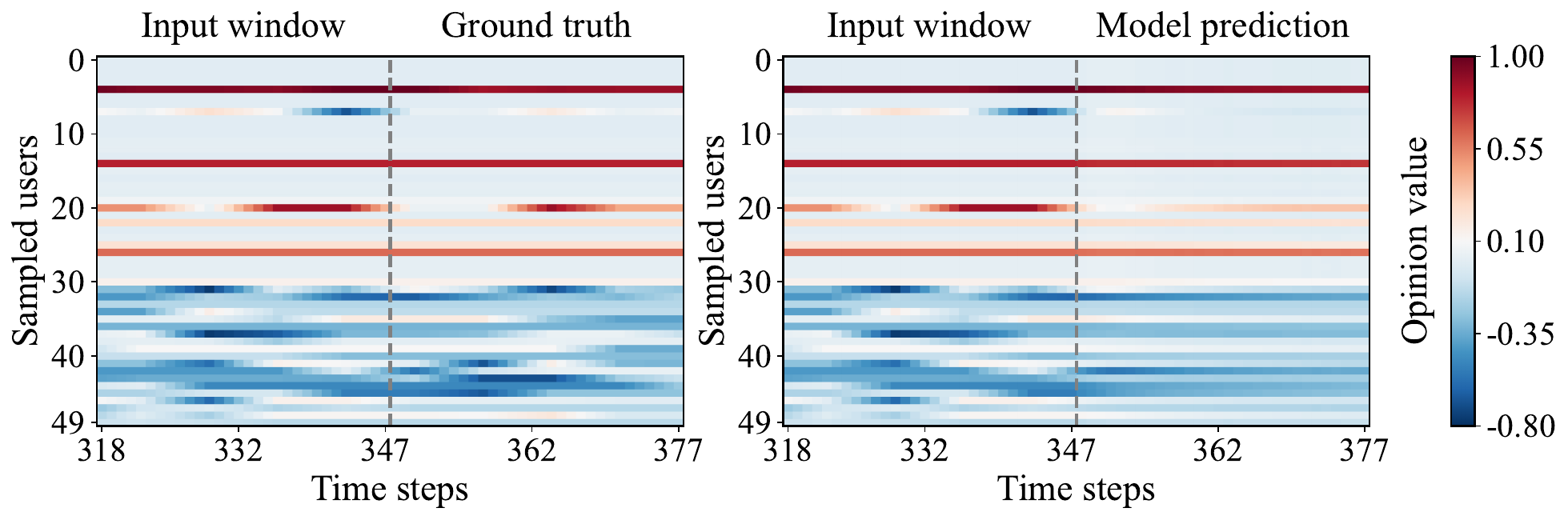} 
        \caption{Visualization of opinions for 50 randomly sampled users.}
    \end{subfigure}
    \begin{subfigure}{0.49\textwidth}
        \centering
        \includegraphics[width=\textwidth]{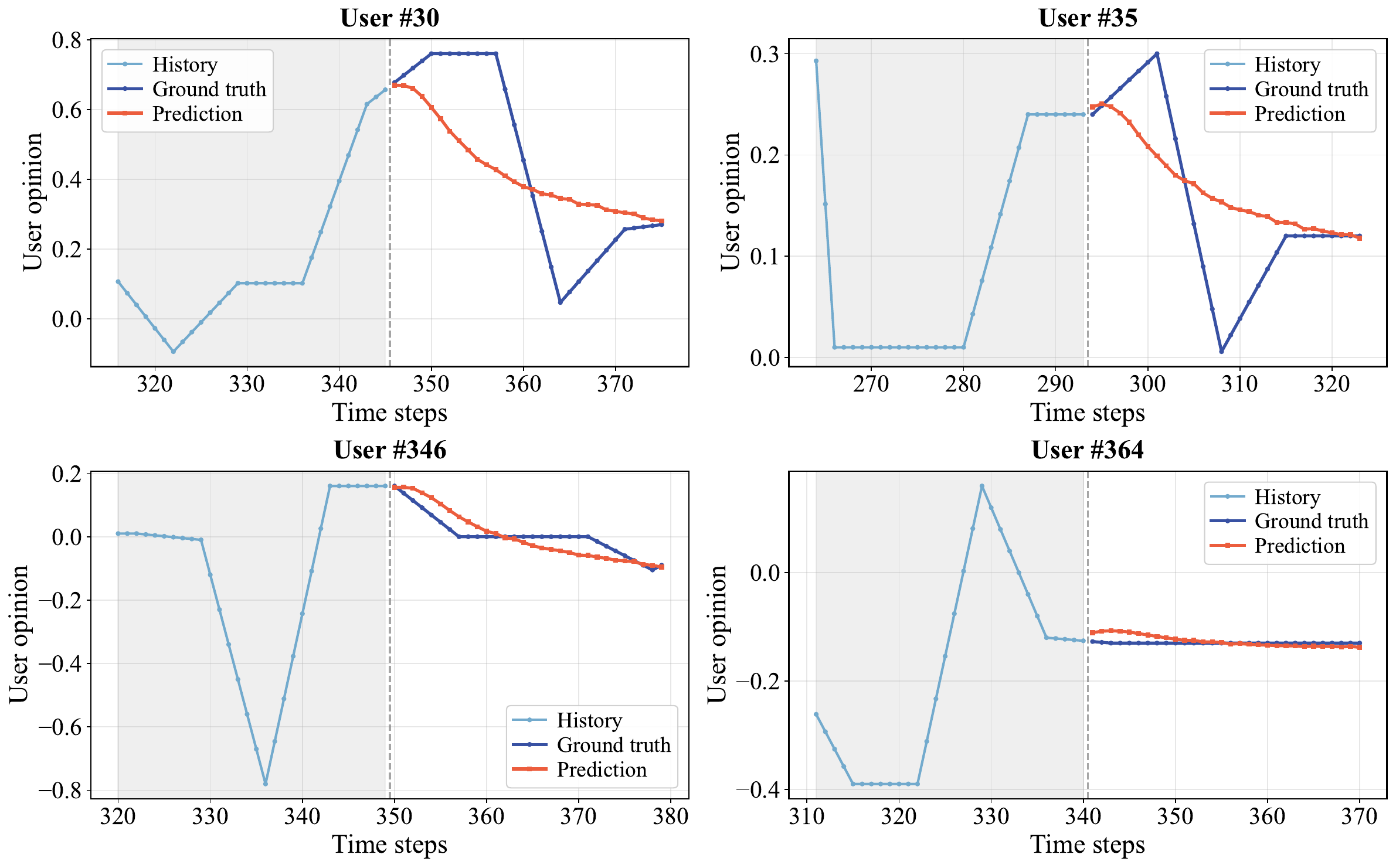} 
        \caption{Visualization of opinions for 4 typical users.}
    \end{subfigure}
    \caption{Case studies on the COVID-19 dataset.} 
    \label{appcs1}
\end{figure}

\begin{figure}[htbp]
    \centering
    \begin{subfigure}{0.49\textwidth}
        \centering
        \includegraphics[width=\textwidth]{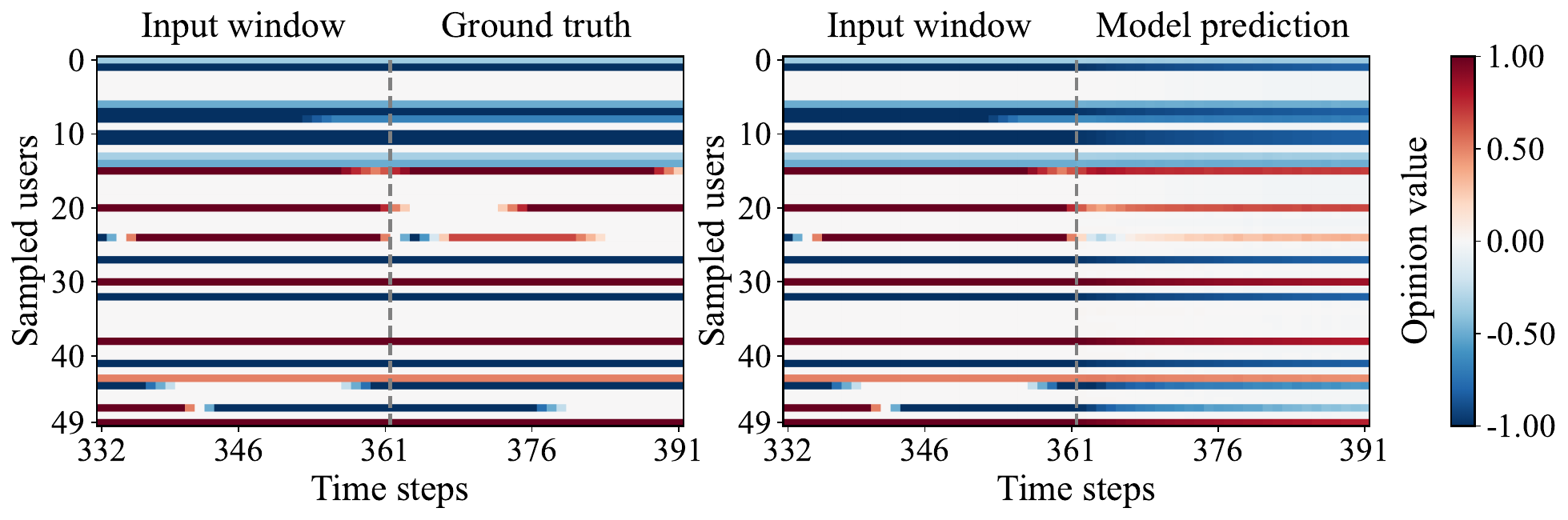} 
        \caption{Visualization of opinions for 50 randomly sampled users.}
    \end{subfigure}
    \begin{subfigure}{0.49\textwidth}
        \centering
        \includegraphics[width=\textwidth]{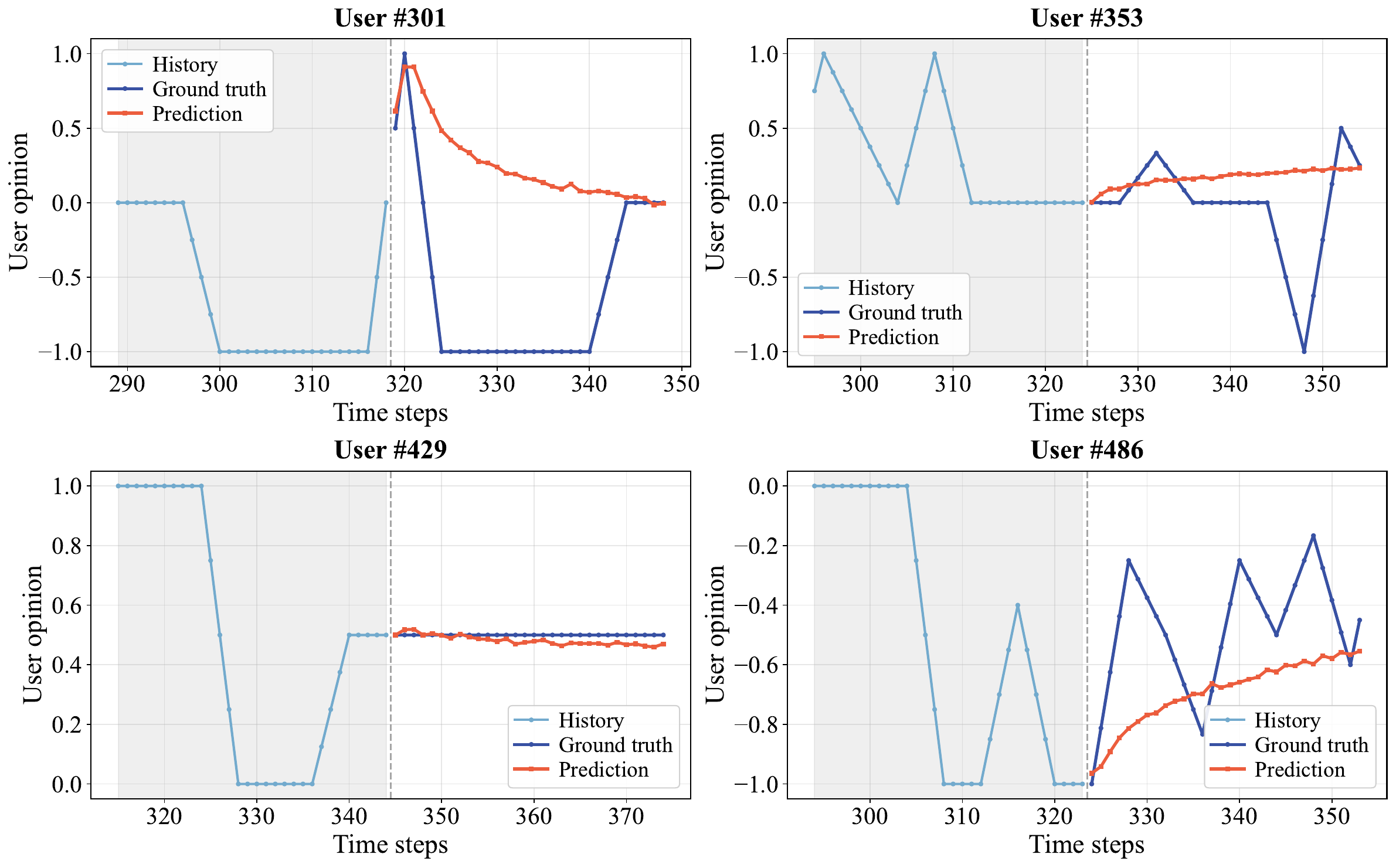} 
        \caption{Visualization of opinions for 4 typical users.}
    \end{subfigure}
    \caption{Case studies on the U.S. Election dataset.} 
    \label{appcs2}
\end{figure}

To complement the case study of Delhi Election presented in the main text, we further evaluate the predictive performance of \textsc{Opinn} on the U.S. Election and COVID-19 datasets, with visualizations provided in Figure~\ref{appcs1} and Figure~\ref{appcs2}.
The findings are consistent with the main text: \textsc{Opinn} can accurately predict the user opinion evolution at both macro and micro levels, even within the inherent noise of real-world social platforms.




\end{document}